\documentclass[]{fairmeta}
\usepackage{amsfonts}
\usepackage{todonotes}
\usepackage{float}
\usepackage{amsmath}
\usepackage{ulem}
\usepackage{wrapfig}

\newif\ifcomment

\commentfalse

\begin{document}

% handle comments in a way to easily enable/disable
% To add a comment command for yourself, either use \pinaforecomment like below
% Or add your custom comment in two places:
% (1) The actual comment below \ifcomment
% (2) A no-op version after \else
\definecolor{lightblue}{HTML}{3cc7ea}
\ifcomment
    \newcommand{\pinaforecomment}[3]{\colorbox{#1}{\parbox{.8\linewidth}{#2: #3}}}
    \newcommand{\todoArti}[1]{\todo[color=cyan!40]{Arti: #1}}
    \newcommand{\todoSrini}[1]{\todo[color=cyan!40]{Srini: #1}}
\else
    \newcommand{\pinaforecomment}[3]{}
    \newcommand{\todoArti}[1]{}
    \newcommand{\todoSrini}[1]{}
\fi

\newcommand{\prcomment}[1]{\pinaforecomment{lightblue}{Pedro}{#1}}
\newcommand{\mike}[1]{\pinaforecomment{green}{Mike}{#1}}
\newcommand{\ari}[1]{\pinaforecomment{purple}{Ari}{#1}}
\newcommand{\margaret}[1]{\pinaforecomment{pink}{Margaret}{#1}}
\newcommand{\ben}[1]{\pinaforecomment{orange}{Ben}{#1}}

\newcommand{\abr}[1]{\textsc{#1}}

% Option "twocolumn" available, but please prioritize single-column

\title{Byte Latent Transformer: Patches Scale Better Than Tokens}

\author[]{Artidoro Pagnoni}
\author[\ddagger]{Ram Pasunuru}
\author[\ddagger]{Pedro Rodriguez}
\author[\ddagger]{John Nguyen}
\author[]{Benjamin Muller}
\author[1,\diamond]{Margaret Li}
\author[\diamond]{Chunting Zhou}
\author[]{Lili Yu}
\author[]{Jason Weston}
\author[]{Luke Zettlemoyer}
\author[]{Gargi Ghosh}
\author[]{Mike Lewis}
\author[\dagger,2,\diamond]{Ari Holtzman}
\author[\dagger]{Srinivasan Iyer}

\affiliation[]{FAIR at Meta}
\affiliation[1]{Paul G. Allen School of Computer Science \& Engineering, University of Washington}
\affiliation[2]{University of Chicago}

\contribution[\ddagger]{Joint second author}
\contribution[\dagger]{Joint last author}
\contribution[\diamond]{Work done at Meta}

\newcommand{\model}{{\textsc BLT}}
\newcommand{\modelbf}{\textbf{BLT}}
\newcommand{\llm}{\abr{llm}}
\newcommand{\llms}{\abr{llm}s}
\newcommand{\flop}{\abr{flop}}
\newcommand{\bpe}{\abr{bpe}}
\newcommand{\absv}[1]{\left | #1 \right |}
\newcommand{\llama}{Llama}

\abstract{
We introduce the Byte Latent Transformer (\model{}), a new byte-level LLM architecture that, for the first time, matches tokenization-based LLM performance at scale with significant improvements in inference efficiency and robustness.
\model{} encodes bytes into dynamically sized patches, which serve as the primary units of computation.
Patches are segmented based on the entropy of the next byte, allocating more compute and model capacity where increased data complexity demands it.
We present the first \flop{} controlled scaling study of byte-level models up to 8B parameters and 4T training bytes. Our results demonstrate the feasibility of scaling models trained on raw bytes without a fixed vocabulary.
Both training and inference efficiency improve due to dynamically selecting long patches when data is predictable, along with qualitative improvements on reasoning and long tail generalization. Overall, for fixed inference costs, \model{} shows significantly better scaling than tokenization-based models, by simultaneously growing both patch and model size. 

}

\date{\today}
\correspondence{artidoro at \email{cs.washington.edu}, sviyer at \email{meta.com}}

% You can add additional metadata fields as follows 
\metadata[Code]{\url{https://github.com/facebookresearch/blt}}
% \metadata[Blogpost]{\url{https://ai.meta.com/blog/?page=1}}

\maketitle

\section{Introduction}
\label{section:intro}

\begin{figure}[t]
    \centering
\centering
    \begin{minipage}{0.5\textwidth}
        \centering
        \includegraphics[width=\linewidth]{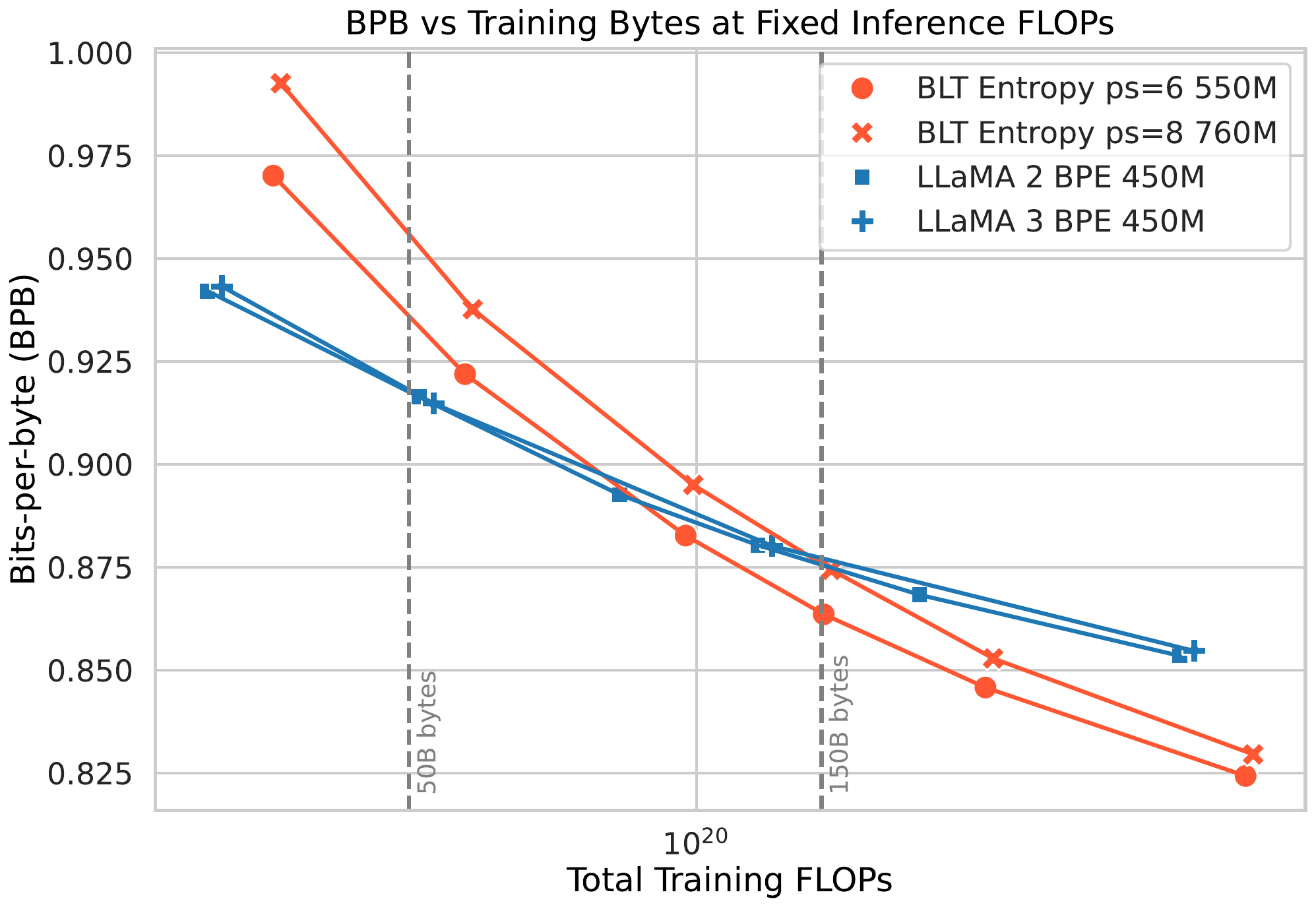}
    \end{minipage}%
    \hfill
    \begin{minipage}{0.5\textwidth}
        \centering
        \includegraphics[width=\linewidth]{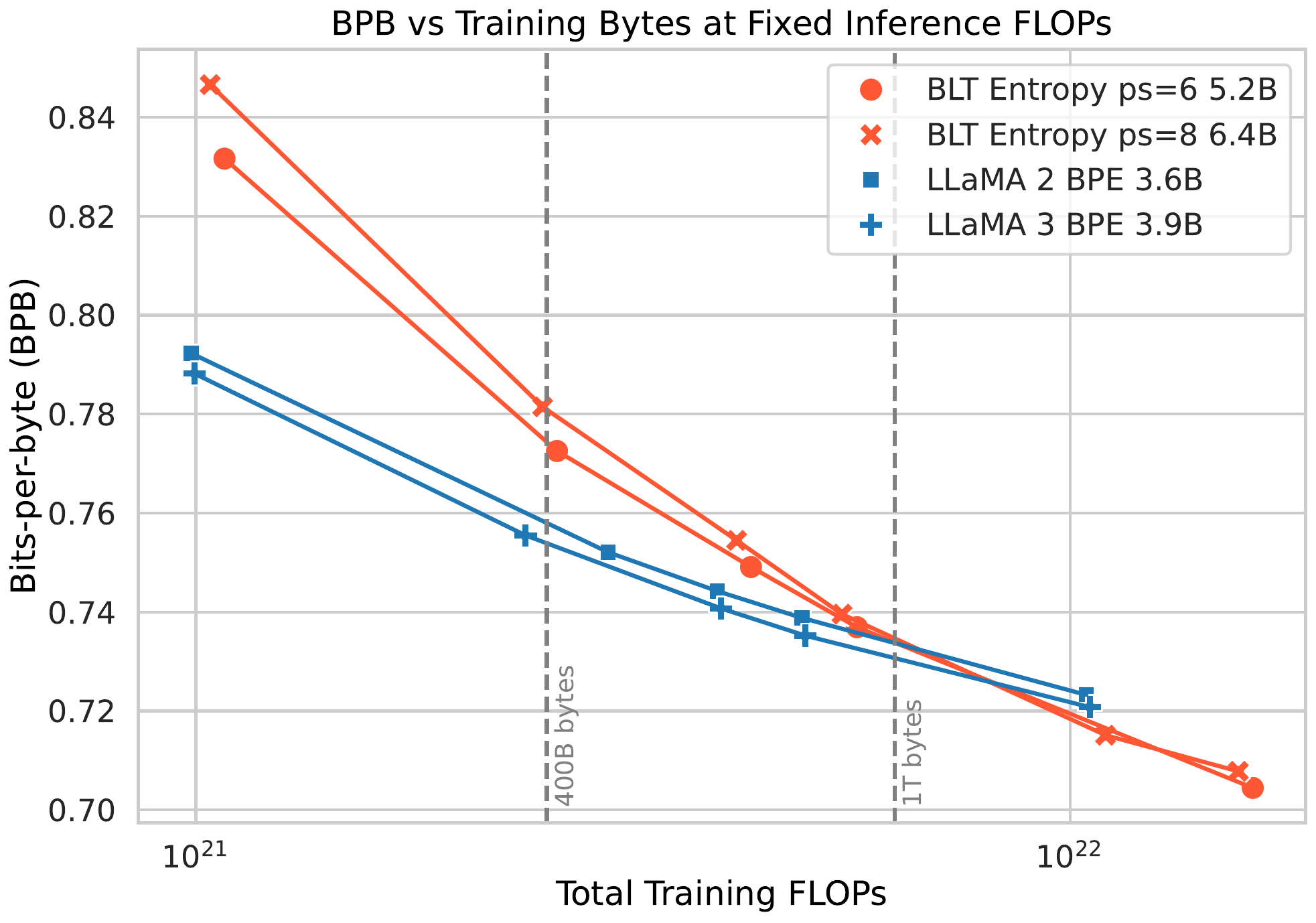}
    \end{minipage}
    \caption{
    Scaling trends for fixed inference \flop{} models (fully) trained with varying training budgets.
    In token-based models, a fixed inference budget determines the model size.
    In contrast, the BLT architecture provides a new scaling axis allowing simultaneous increases in model and patch size while keeping the same training and inference budget.
    \model{} patch-size (ps) 6 and 8 models quickly overtake scaling trends of \abr{bpe} \llama{}~2 and 3. Moving to the larger inference budget makes the larger patch size 8 model more desirable sooner. Both BPE compute-optimal point and crossover point are indicated with vertical lines.
    }
    \label{fig:fixed_inference_scaling}
\end{figure}
\begin{figure}[t]
    \centering
    \includegraphics[width=0.9\textwidth]{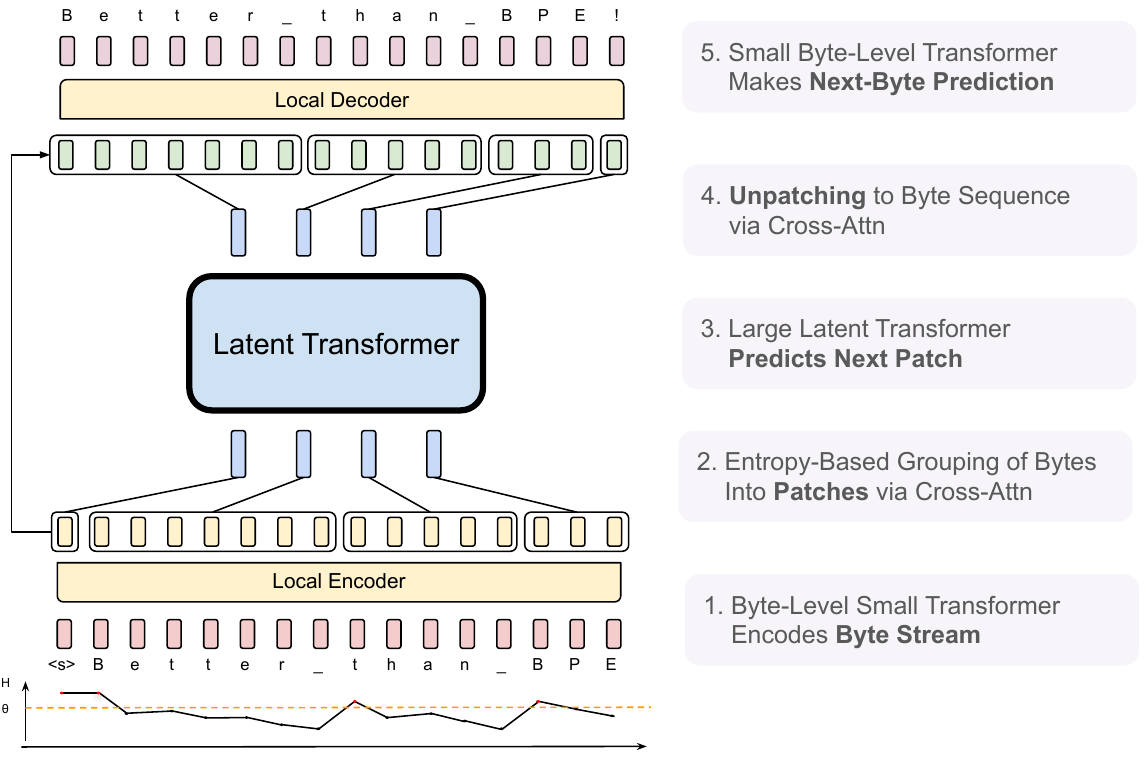}
    \caption{
    \model{} comprises three modules, a lightweight \textit{Local Encoder} that encodes input bytes into patch representations, a computationally expensive Latent Transformer over patch representations, and a lightweight \textit{Local Decoder} to decode the next patch of bytes. \model{} incorporates byte $n$-gram embeddings and a cross-attention mechanism to maximize information flow between the Latent Transformer and the byte-level modules~(\autoref{fig:crossattn}). Unlike fixed-vocabulary tokenization, \model{} dynamically groups bytes into patches preserving access to the byte-level information.
    }
    \label{fig:arch_high_level}
\end{figure}

We introduce the Byte Latent Transformer~(\modelbf{}), a tokenizer-free architecture that learns from raw byte data and, for the first time, matches the performance of tokenization-based models at scale, with significant improvements in efficiency and robustness (\S\ref{sec:robustness}).
Existing large language models (\llms{}) are trained almost entirely end-to-end, except for tokenization---a heuristic pre-processing step that groups bytes into a static set of tokens.
Such tokens bias how a string is compressed, leading to shortcomings such as domain/modality sensitivity~\citep{dagangetting}, sensitivity to input noise~(\S\ref{sec:robustness}), a lack of orthographic knowledge~\citep{edman2024cute}, and multilingual inequity~\citep{liang2023xlm,petrov2024language,limisiewicz2024myte}. 

Tokenization has previously been essential because directly training \llm{}s on bytes is prohibitively costly at scale due to long sequence lengths~\citep{xue2022byt5}.
Prior works mitigate this by employing more efficient self-attention~\citep{el2020characterbert, clark2022canine} or attention-free architectures~\citep{wang2024mambabyte}~(\S\ref{section:related_work}). However, this primarily helps train \textit{small models}.
At scale, the computational cost of a Transformer is dominated by large feed-forward network layers that run on every byte, not the cost of the attention mechanism.

To efficiently allocate compute, we propose a dynamic, learnable method for grouping bytes into \textit{patches}~(\S\ref{section:patching}) and a new model architecture that mixes byte and patch information.
Unlike tokenization, \model{} has no fixed vocabulary for patches.
Arbitrary groups of bytes are mapped to latent patch representations via light-weight learned encoder and decoder modules.
We show that this results in \textit{more} efficient allocation of compute than tokenization-based models.
\todoArti{Should we try to cite some byte-level work here? space-byte, megabyte, dynamic token-pooling, ByT5, MyT5. In general intro lacks citations. [[Luke: yes!]]}

Tokenization-based \llm{}s allocate the same amount of compute to every token. This trades efficiency for performance, since tokens are induced with compression heuristics that are not always correlated with the complexity of predictions. Central to our architecture is the idea that models should dynamically allocate compute where it is needed. For example, a large transformer is not needed to predict the ending of most words, since these are comparably easy, low-entropy decisions compared to choosing the first word of a new sentence.
This is reflected in \model{}'s architecture~(\S\ref{section:architecture}) where there are three transformer blocks: two small byte-level \textit{local models} and a large global \textit{latent transformer}~(\autoref{fig:arch_high_level}).
To determine how to group bytes into patches and therefore how to dynamically allocate compute, \model{}
segments data based on the entropy of the next-byte prediction creating contextualized groupings of bytes with relatively uniform information density.

\todoArti{We should discuss the importance of inference time scaling (o1 style)}

We present the first \flop{}-controlled scaling study of byte-level models up to 8B parameters and 4T training bytes, showing that we can train a model end-to-end at scale from bytes without fixed-vocabulary tokenization. 
Overall, \model{} matches training \flop{}-controlled performance\footnote{We calculate the computational cost of a model by counting the number of Floating Point OPerations (\flop{}s) needed.} of \llama{} 3 while using up to 50\% fewer \flop{}s at inference~(\S\ref{section:scaling}).
We also show that directly working with raw bytes provides significant improvements in modeling the long-tail of the data. \model{} models are more robust than tokenizer-based models to noisy inputs and display enhanced character level understanding abilities demonstrated on orthographic knowledge, phonology, and low-resource machine translation tasks~(\S\ref{sec:robustness}).
Finally, with \model{} models, we can simultaneously increase model size and patch size  while maintaining the same inference \flop{} budget. Longer patch sizes, on average, save compute which can be reallocated to grow the size of the global latent transformer, because it is run less often. We conduct inference-\flop{} controlled scaling experiments~(\autoref{fig:fixed_inference_scaling}), and observe significantly better scaling trends than with tokenization-based architectures. 

\todoArti{We might want to front load a bit more this result? And maybe expand on it a little more}

In summary, this paper makes the following contributions:
1) We introduce \model{}, a byte latent \llm{} architecture that dynamically allocates compute to improve \flop{} efficiency,
2) We show that we achieve training \flop{}-controlled parity with \llama{} 3 up to 8B scale while having the option to trade minor losses in evaluation metrics for \flop{} efficiency gains of up to 50\%, 3) \model{} models unlock a new dimension for scaling \llm{}s, where model size can now be scaled while maintaining a fixed-inference budget, 4) We demonstrate the improved robustness of \model{} models to input noise and their awareness of sub-word aspects of input data that token-based \llms{} miss.
We release the training and inference code for \model{}  at~\url{https://github.com/facebookresearch/blt}.

\section{Patching: From Individual Bytes to Groups of Bytes}
\label{section:patching}
\todoArti{Should we frontload the discussion of patching vs tokenization? it's currently in the bpe section at the end}
\todoArti{I think we should start by saying that a patch still has access to the underlying byte information}
\begin{figure}[t]
    \centering
    \includegraphics[width=\textwidth]{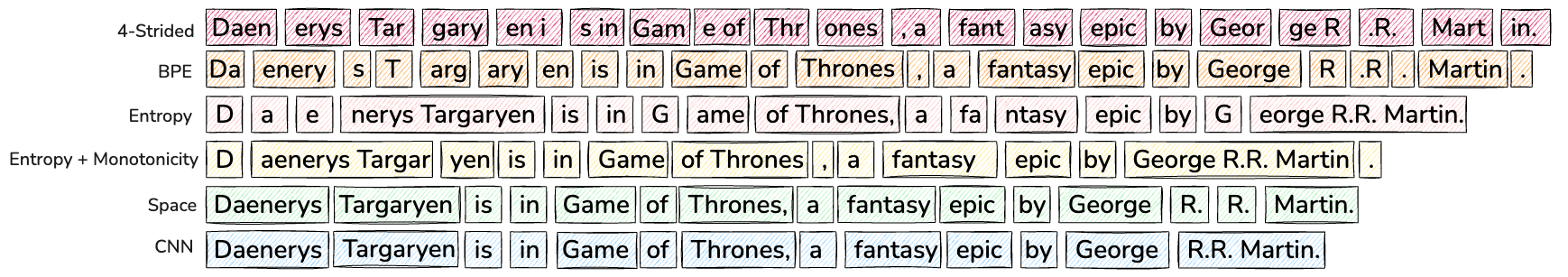}
    \caption{
    Patching schemes group bytes in different ways, each leading to a different number of resulting patches.
    Since each patch is processed using a large transformer step, the number of patches directly determines the bulk of the compute expended in terms of \flop{}s.
    These schemes group bytes into patches by (a) striding every four bytes~(\S\ref{section:static-patch}) as in MegaByte~\citep{yu2023megabyte}, (b) tokenizing with Byte-Pair Encoding (\bpe{}), in this case the \llama{}-3~\citep{dubey2024llama} tokenizer, (c \& d) entropy-based patching as in this work~(\S\ref{section:dyn-patch}), (e) patching on space-bytes~\citep{slagle2024spacebyte}, (f) and patching on entropy using a small CNN byte-level model with 2-byte context.}
    \label{fig:patching_types}
\end{figure}

Segmenting bytes into \textit{patches} allows \model{} to dynamically allocate compute based on context.
Figure~\ref{fig:patching_types} shows several different methods for segmenting bytes into patches.
Formally, a patching function $f_p$ segments a sequence of bytes $\pmb{x}=\{x_i,|i=1,\ldots n\}$ of length $n$ into a sequence of $m < n$ patches $\pmb{p}=\{p_j|j=1,\ldots,m\}$ by mapping each $x_i$ to the set \{0,1\} where 1 indicates the start of a new patch. 
For both token-based and patch-based models, the computational cost of processing data is primarily determined by the number of steps executed by the main Transformer. In \model{}, this is the number of patches needed to encode the data with a given patching function. 
Consequently, the average size of a patch, or simply \textit{patch size}, is the main factor for determining the cost of processing data during both training and inference with a given patching function~(\S\ref{section:flops}).
Next, we introduce three patching functions: patching with a fixed number of bytes per patch~(\S\ref{section:static-patch}), whitespace patching~(\S\ref{section:white-patch}), and dynamically patching with entropies from a small byte \abr{lm}~(\S\ref{section:dyn-patch}). Finally, we discuss incremental patching and how tokenization is different from patching~(\S\ref{section:bpe-patch}).

\subsection{Strided Patching Every K Bytes}
\label{section:static-patch}

Perhaps the most straightforward way to group bytes is into patches of fixed size $k$ as done in MegaByte~\citep{yu2023megabyte}.
The fixed stride is easy to implement for training and inference, provides a straightforward mechanism for changing the average patch size, and therefore makes it easy to control the \flop{} cost.
However, this patching function comes with significant downsides.
First, compute is not dynamically allocated to where it is needed most: one could be either wasting a transformer step $j$ if only predicting whitespace in code, or not allocating sufficient compute for bytes dense with information such as math.
Second, this leads to inconsistent and non-contextual patching of similar byte sequences, such as the same word being split differently.

\subsection{Space Patching}
\label{section:white-patch}

\citet{slagle2024spacebyte} proposes a simple yet effective
improvement over strided patching that creates new patches after any space-like bytes\footnote{
    Space-like bytes are defined as any byte that is not a latin character, digit, or \abr{utf}-8 continuation byte.
    In addition, each patch must contain at least one non space-like byte.
}
which are natural boundaries for linguistic units in many languages. 
In Space patching, a latent transformer step (i.e., more \flop{}s) is allocated to model every word.
This ensures words are patched in the same way across sequences and that flops are allocated for hard predictions which often follow spaces. For example, predicting the first byte of the answer to the question ``Who composed the Magic Flute?\uline{\hspace{.6em}}'' is much harder than predicting the remaining bytes after ``M'' since the first character significantly reduces the number of likely choices, making the completion ``Mozart'' comparatively easy to predict.
However, space patching cannot gracefully handle all languages and domains, and most importantly cannot vary the patch size.
Next, we introduce a new patching method that uses the insight that the first bytes in words are typically most difficult to predict, but that provides a natural mechanism for controlling patch size.

\subsection{Entropy Patching: Using Next-Byte Entropies from a Small Byte LM}
\label{section:dyn-patch}

Rather than relying on a rule-based heuristic such as whitespace, we instead take a data-driven approach to identify high uncertainty next-byte predictions.
We introduce \textit{entropy patching}, which uses entropy estimates to derive patch boundaries.

We train a small byte-level auto-regressive language model on the training data for \model{} and compute next byte entropies under the LM distribution $p_{e}$ over the byte vocabulary $\mathcal{V}$:
\begin{equation}
H(x_i) = \sum_{v \in \mathcal{V}} p_{e}(x_i=v|\pmb{x}_{<i}) \log p_{e}(x_i=v|\pmb{x}_{<i})
\end{equation}

\begin{figure}[t]
    \centering
    \includegraphics[width=\textwidth]{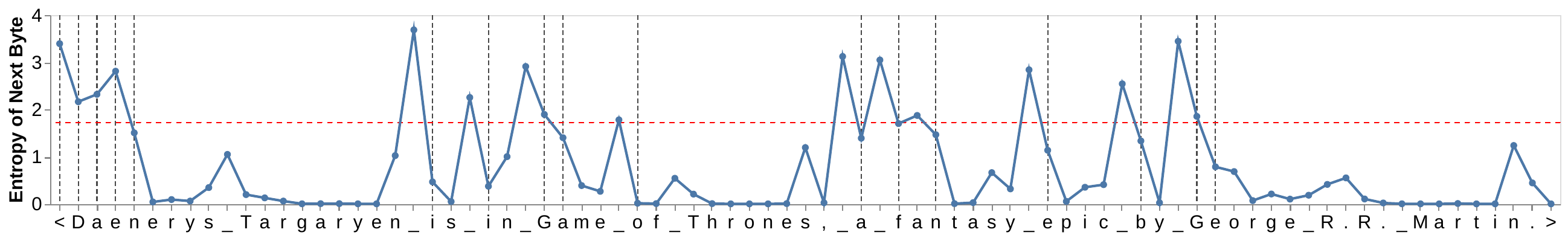}
    \caption{
    This figure plots the entropy $H(x_i)$ of each byte in ``Daenerys Targeryen is in Game of Thrones, a fantasy epic by George R.R. Martin.'' with spaces shown as underscores.
    Patches end when $H(x_i)$ exceeds the global threshold $\theta_g$, shown as a red horizontal line.
    The start of new patches are shown with vertical gray lines.
    For example, the entropies of ``G'' and ``e'' in ``George R.R. Martin'' exceed $\theta_g$, so ``G'' is the start of a single byte patch and ``e'' of a larger patch extending to the end of the named entity as the entropy $H(x_i)$ stays low, resulting in no additional patches.
    }
    \label{fig:patching}
\end{figure}

We experiment with two methods to identify patch boundaries given entropies $H(x_i)$. The first, finds points above a global entropy threshold, as illustrated in~\autoref{fig:patching}. The second, identifies points that are high relative to the previous entropy. The second approach can also be interpreted as identifying points that break approximate monotonically decreasing entropy withing the patch.
\begin{align*}
    \text{Global Constraint}\hspace{1cm}H(x_t)& > \theta_g\\
    \text{Approx. Monotonic Constraint}\hspace{1cm}H(x_t)& - H(x_{t-1}) > \theta_r
\end{align*}
\todoArti{Should we explain when these patching schemes might be better? i.e. monotonicity should help achieve more consistent patching bc of decay of entropy with more context, but in practice global works well}
Patch boundaries are identified during a lightweight preprocessing step executed during dataloading. This is different from~\citet{nawrot-etal-2023-efficient} where classifier is trained to predict entropy-based patch boundaries.
In our experiments~(\S\ref{section:experiments}), we compare these two methods for distinguishing between low and high entropy bytes. \todoArti{spoiler sentence}

\subsection{The Byte-Pair Encoding (BPE) Tokenizer and Incremental Patching}
\label{section:incremental-patching}
\label{section:bpe-patch}

Many modern \llms{}, including our baseline \llama{} 3, use a subword tokenizer like \bpe{}~\citep{gage1994new, sennrich-etal-2016-neural}. 
We use ``tokens'' to refer to byte-groups drawn from a \textit{finite} vocabulary determined prior to training as opposed to ``patches'' which refer to dynamically grouped sequences without a fixed vocabulary. 
A critical difference between patches and tokens is that with tokens, the model has no direct access to the underlying byte features. 

A crucial improvement of \model{} over tokenization-based models is that redefines the trade off between the vocabulary size and compute. In standard \llms{}, increasing the size of the vocabulary means larger tokens on average and therefore fewer steps for the model but also larger output dimension for the final projection layer of the model. This trade off effectively leaves little room for tokenization based approaches to achieve significant variations in token size and inference cost. For example, \llama{} 3 increases the average token size from  3.7 to 4.4 bytes at the cost of increasing the size of its embedding table 4x compared to \llama{} 2.

When generating, \model{} needs to decide whether the current step in the byte sequence is at a patch boundary or not as this determines whether more compute is invoked via the Latent Transformer. This decision needs to occur independently of the rest of the sequence which has yet to be generated. Thus patching cannot assume access to future bytes in order to choose how to segment the byte sequence. Formally, a patching scheme $f_p$ satisfies the property of incremental patching if it satisfies:
$$f_p(\pmb{x}_{<i}) = f_p(\pmb{x})_{<i}$$
\bpe{} is not an incremental patching scheme as the same prefix can be tokenized differently depending on the continuation sequence, and therefore does not satisfy the property above\footnote{Using a special delimiter token to indicate patch boundaries can turn \bpe{} into an incremental patching scheme but increases the byte-sequence length.}.\todoArti{Can we say something like: We argue that satisfying this property helps improve the robustness of a model. }

\section{\model{} Architecture}
\label{section:architecture}
\model{} is composed of a large global autoregressive language model that operates on patch representations, along with two smaller local models that encode sequences of bytes into patches and decode patch representations back into bytes~(\autoref{fig:arch_high_level}).

\subsection{Latent Global Transformer Model}

\textit{The Latent Global Transformer} is an autoregressive transformer model $\mathcal{G}$ with $l_{\mathcal{G}}$ layers, which maps a sequence of latent input patch representations, $p_j$ into a sequence of output patch representations, $o_j$. Throughout the paper, we use the subscript $j$ to denote patches and $i$ to denote bytes. The global model uses a block-causal attention mask~\citep{dubey2024llama}, which restricts attention to be up to and including the current patch  within the current document. This model consumes the bulk of the \flop{}s during pre-training as well as inference, and thus, choosing when to invoke it allows us to control and vary the amount of compute expended for different portions of the input and output as a function of input/output complexity.

\subsection{Local Encoder}
\label{section:local}

\begin{figure}[t]
    \centering
    \includegraphics[width=\textwidth]{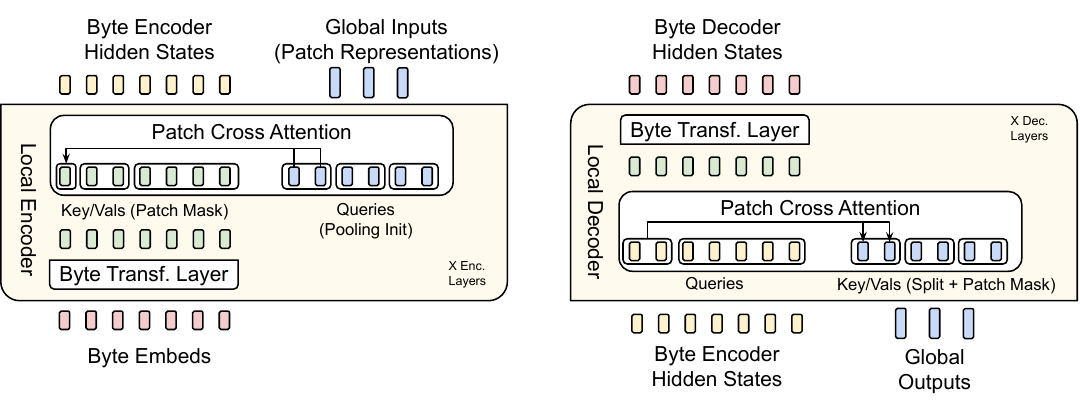}
    \caption{The local encoder uses a cross-attention block with patch representations as queries, and byte representations as keys/values to encode byte representations into patch representations. The local decoder uses a similar block but with the roles reversed i.e. byte representations are now the queries and patch representations are the keys/values. Here we use Cross-Attn $k=2$.
    }
    \label{fig:crossattn}
\end{figure}

\textit{The Local Encoder Model}, denoted by $\mathcal{E}$, is a lightweight transformer-based model with $l_{\mathcal{E}} << l_{\mathcal{G}}$ layers, whose main role is to efficiently map a sequence of input bytes $b_i$, into expressive patch representations, $p_j$.
A primary departure from the transformer architecture is the addition of a cross-attention layer after each transformer layer, whose function is to pool byte representations into patch representations~(\autoref{fig:crossattn}).
First, the input sequence of bytes, $b_i$, are embedded using a $\mathbb{R}^{256 \times h_{\mathcal{E}}}$ matrix, denoted as $x_i$.
These embeddings are then optionally augmented with additional information in the form of hash-embeddings~(\S\ref{sec:hash-emb}).
A series of alternating transformer and cross-attention layers~(\S\ref{sec:enc-cross-attn}) then transform these representations into patch representations, $p_i$ that are processed by the global transformer, $\mathcal{G}$.
The transformer layers use a \textit{local block causal} attention mask; each byte attends to a fixed window of $w_{\mathcal{E}}$ preceding bytes that in general can cross the dynamic patch boundaries but can not cross document boundaries.
The following subsections describe details about the embeddings and the cross-attention block.

\subsubsection{Encoder Hash n-gram Embeddings}
\label{sec:ngram-emb}
\label{sec:hash-emb}
A key component in creating robust, expressive representations at each step $i$ is to incorporate information about the preceding bytes.
In \model{}, we achieve this by modeling both the byte $b_i$ individually \textit{and} as part of a byte n-gram.
For each step $i$, we first construct byte-grams
\begin{align}
    g_{i,n}=\{b_{i-n + 1},\ldots, b_i\}
\end{align}
for each byte position $i$ and $n$ from three to eight.\footnote{
We omit byte-grams of size $n$ or more when $i<n$.
}

We then introduce hash $n$-gram embeddings, that map all byte $n$-grams via a hash function to an index in an embedding table $E_{n}^{hash}$ with a fixed size, for each size $n \in\{3,4,5,6,7,8\}$~\citep{bai2010learning}.
The resulting embedding is then added to the embedding of the byte before being normalized and passed as input to the local encoder model.
We calculate the augmented embedding
\begin{align}
e_i &= x_i + \sum_{n = 3,...,8}E_{n}^{hash}(\text{Hash}(g_{i,n})) \\
\text{where, Hash}(g_{i,n}) &= \text{RollPolyHash}(g_{i,n}) \% |E_{n}^{hash}|
\end{align}
We normalize $e_i$ by the number of $n$-grams sizes plus one and use RollPolyHash as defined in Appendix~\ref{appendix:rollpolyhash}.
In Section~\ref{section:ablations}, we ablate the effects of $n$-gram hash embeddings with different values for $n$ and embedding table size on \flop{}-controlled scaling law trends.
In addition to hash $n$-gram embeddings, we also experimented with frequency based $n$-gram embeddings, and we provide details of this exploration in Appendix~\ref{appendix:ngrams}.

\subsubsection{Encoder Multi-Headed Cross-Attention}
\label{sec:enc-cross-attn}
We closely follow the input cross-attention module of the Perceiver architecture~\citep{jaegle2021perceiver}, with the main difference being that latent representations correspond to variable patch representations as opposed to a fixed set of latent representations~(\autoref{fig:crossattn}), and only attend to the bytes that make up the respective patch.
The module comprises a query vector, corresponding to each patch $p_j$, which is initialized by pooling the byte representations corresponding to patch $p_j$, followed by a linear projection, $\mathcal{E}_{C} \in \mathbb{R}^{h_{\mathcal{E}} \times (h_{\mathcal{E}} \times U_{\mathcal{E}})}$, where $U_{\mathcal{E}}$ is the number of encoder cross-attention heads.
Formally, if we let $f_{\text{bytes}}(p_j)$ denote the sequence of bytes corresponding to patch, $p_j$, then we calculate
\begin{align}
P_{0,j} &= \mathcal{E}_{C}(f_\text{bytes}((p_j)), f~ \text{is a pooling function} \\
P_l &= P_{l-1} + W_o\left(\text{softmax}\left(\frac{QK^T}{\sqrt{d_k}}\right)V\right) \\
\text{where } Q_j &= W_q(P_{l-1,j}), K_i = W_k(h_{l-1,i}), V_i = W_v(h_{l-1,i}) \\
h_l &= \text{Encoder-Transformer-Layer}_l(h_{l-1})
\end{align}
where $P \in \mathbb{R}^{n_p \times h_{\mathcal{G}}}$ represents $n_p$ patch representations to be processed by the global model, which is initialized by pooling together the byte embeddings $e_i$ corresponding to each patch $p_j$.
$W_q$, $W_k$, $W_v$ and $W_o$ are the projections corresponding to the queries, keys, values, and output where the keys and values are projections of byte representations $h_i$ from the previous layer ($e_i$ for the first layer).
We use a masking strategy specific to patching where each query $Q_j$ only attends to the keys and values that correspond to the bytes in patch $j$. Because we use multi-headed attention over $Q, K$ and $V$ and patch representations are typically of larger dimension ($h_{\mathcal{G}}$) than $h_{\mathcal{E}}$, we maintain $P_l$ as multiple heads of dimension $h_{\mathcal{E}}$ when doing cross-attention, and later, concat these representations into $h_{\mathcal{G}}$ dimensions.
Additionally, we use a pre-LayerNorm on the queries, keys and values and no positional embeddings are used in this cross-attention module.
Finally, we use a residual connection around the cross-attention block. 

\subsection{Local Decoder} 

Similar to the local encoder, the local decoder  $\mathcal{D}$ is a lightweight transformer-based model with $l_{\mathcal{D}} << l_{\mathcal{G}}$ layers, that decodes a sequence of global patch representations $o_j$, into raw bytes, $y_i$. The local decoder predicts a sequence of raw bytes, as a function of previously decoded bytes, and thus, takes as input the hidden representations produced by the local encoder for the byte-sequence.
It applies a series of $l_{\mathcal{D}}$ alternating layers of cross attention and transformer layers. The cross-attention layer in the decoder is applied before the transformer layer to first create byte representations from the patch representations, and the local decoder transformer layer operates on the resulting byte sequence. 

\subsubsection{Decoder Multi-headed Cross-Attention} 

In the decoder cross-attention, the roles of the queries and key/values are interchanged i.e. the byte-representations are now the queries, and the patch representations are now the key/values. The initial byte-representations for the cross-attention are initialized as the byte embeddings from the last encoder layer i.e. $h_{l_{\mathcal{E}}}$. The subsequent byte-representations for layer $l$, $d_{l,i}$ are computed as:

\begin{align}
D_0 &= h_{l_{\mathcal{E}}} \\
B_l &= D_{l-1} + W_o\left(\text{softmax}\left(\frac{QK^T}{\sqrt{d_k}}\right)V\right), \\
  &\text{where } Q_i = W_q(d_{l-1,i}), K_i = W_k(\mathcal{D}_C(o_j)), V_i = W_v(\mathcal{D}_C(o_j)) \\
  D_l &= \text{Decoder-Transformer-layer}_l(B_l)
\end{align}

where once again, $W_k, W_v$ are key/value projection matrices that operate on a linear transformation and split operation $\mathcal{D}_C$, applied to the final patch representations $o_j$ from the global model, $W_q$ is a query projection matrices operating on byte representations $d_{l-1}$ from the previous decoder transformer layer (or $h_{l_{\mathcal{E}}}$ for the first layer), and $W_o$ is the output projection matrix, thus making $B \in \mathbb{R}^{h_{\mathcal{D}} \times n_b}$, where $n_b$ is the number of output bytes. The next decoder representations $D_l$ are computed using a decoder transformer layer on the output of the cross-attention block, $B$. As in the local encoder cross-attention, we use multiple heads in the attention, use pre LayerNorms, no positional embeddings, and a residual connection around the cross-attention module. 

\section{Experimental Setup}
\label{section:experiments}
We carefully design controlled experiments to compare \model{} with tokenization based models with particular attention to not give \model{} any advantages from possibly using longer sequence contexts.

\subsection{Pre-training Datasets}
All model scales that we experiment in this paper are pre-trained on two datasets: 1) The \llama{} 2 dataset~\citep{touvron2023llama}, which comprises 2 trillion tokens collected from a variety of publicly available sources, which are subsequently cleaned and filtered to improve quality; and 2) \model{}-1T: A new dataset with 1 trillion tokens gathered from various public sources, and also including a subset of the pre-training data released by Datacomp-LM~\citep{li2024datacomp}. The former is used for scaling law experiments on optimal number of tokens as determined by~\cite{dubey2024llama} to determine the best architectural choices for \model{}, while the latter is used for a complete pre-training run to compare with \llama{} 3 on downstream tasks. Neither of these datasets include any data gathered from Meta products or services.
Furthermore, for baseline experiments for tokenizer-based models, we use the \llama{} 3 tokenizer with a vocabulary size of 128K tokens, which produced stronger baseline performance that the \llama{} 2 tokenizer in our experiments.

\subsection{Entropy Model}
The entropy model in our experiments is a byte level language model trained on the same training distribution as the full \model{} model. Unless otherwise mentioned, we use a transformer with 100M parameters, 14 layers, and a hidden dimensionality of 512, and sliding window attention of 512 bytes. The remaining hyperparameters are the same as in our local and global transformers. We  experimented with different model sizes, receptive fields, and architectures as discussed in \autoref{section:ablations}. In particular, when the receptive field of the model is small enough, the trained entropy model can be encoded in an efficient lookup table.

\subsection{Entropy Threshold and Equalizing Context Length}
For models using entropy-based patching, we estimate a patching threshold that achieves a desired average \textit{patch size} on the pretraining data mix.  In \model{}, unlike with tokenization, the \textit{patch size} can be arbitrarily chosen having significant implications on the context size used by the model.
To maintain the same average context length and avoid giving larger patch sizes unfair advantage, we ensure that the number of bytes in each batch remains constant in expectation. This means that we reduce the sequence length of models with larger patch sizes. On \llama{} 2 data, we use a 8k byte context while on the \model{}-1T dataset we increase the context to 16k bytes on average while maintaining the same batch size of 16M bytes on average.

While the average batch size is constant, when loading batches of data, dynamic patching methods yield different ratios of bytes to patches.
For efficiency reasons, our implementation of \model{} training packs batches of patches to avoid padding steps in the more expensive latent transformer. This ensures that every batch has the same number of patches. During training we pad and possibly truncate byte sequences to 12k and 24k bytes respectively for \llama{} 2 and \model{}-1T datasets, to avoid memory spikes from sequences with unusually large patches.

\subsection{Entropy Model Context}
\label{section:entropy-context}
Empirically, we find that using entropy patching yields progressively larger patches in structured content like multiple choice tasks (see patching on an MMLU example in \autoref{fig:mmlu_patching}) which are often very repetitive. These variations are caused by lower entropy on the repeated content found in the entropy model context. So for the large scale run of \model{}-Entropy with patch size 4.5, we reset the entropy context with new lines and use approximate monontonicity constraint as it suffers less from "entropy drift" from changes in context length. This change only affects how we compute entropies, but we still follow the same procedure to identify the value of the entropy threshold.

\subsection{FLOPs Estimation} 
\label{section:flops}

We largely follow the equations for computation of transformer \flop{}s from Chinchilla~\citep{hoffmann2022training} comprising \flop{}s for the feed-forward layers, \abr{qkvo} projections in the self-attention layer, and computation of attention and output projection.
A notable difference is that we assume the input embedding layer is implemented as an efficient lookup instead of a dense matrix multiplication, therefore becoming a 0-\flop{} operation.
Following previous work, we estimate that the backwards pass has twice the number of \flop{}s as the forward pass.

To compute \flop{}s \textit{per byte} for \model{} models, we add up the \flop{}s for the local encoder transformer, the global latent transformer, and the local decoder transformer, together with the cross attention blocks in the encoder and the decoder:

\begin{align}
    \text{FL}_{\text{\model{}}} &= \text{Transf. FL}(h_{\mathcal{G}}, l_{\mathcal{G}}, m=n_{ctx}/n_p,V=0) / n_p\\
   &+\text{Transf. FL}(h_{\mathcal{E}}, l_{\mathcal{E}}, m=w_{\mathcal{E}}, V=0) \\
   &+ \text{Transf. FL}(h_{\mathcal{D}}, l_{\mathcal{D}}, m=w_{\mathcal{D}}, V=256) \\ 
    &+ \text{Cross Attn. FL}(h_{\mathcal{E}}, l_{\mathcal{E}}, m=n_p, r=n_p/k) \times k/n_p \\ 
   &+ \text{Cross Attn. FL}(h_{\mathcal{D}}, l_{\mathcal{D}}, m=k, r=k/n_p) 
\end{align}
where $n_{ctx}$ is the sequence length in bytes, $n_p$ is the patch size, $r$ is the ratio of queries to key/values, $k$ is the ratio of patch-dimension to byte-dimension i.e. the number of local model splits that concatenate to form a global model representation ($k=2$ in \autoref{fig:crossattn}). $V$ corresponds to the vocabulary size for the output projection, which is only used in the local decoder.
Depending on whether a module is applied on the byte or patch sequence, the attention uses a different context length, $m$. We modify the attention \flop{}s accordingly for each component.
The exact equations for \flop{}s computation for Transformer-FLOPs and Cross-Attention FLOPs are provided in Appendix~\ref{section:flops-appendix}.

\subsection{Bits-Per-Byte Estimation}
Perplexity only makes sense in the context of a fixed tokenizer as it is a measure of the uncertainty for each token. When comparing byte and token-level models, following previous work~\citep{xue2022byt5, yu2023megabyte, wang2024mambabyte}, we instead report Bits-Per-Byte (BPB), a tokenizer independent version of perplexity. Specifically:
\begin{align}
    \text{BPB}(x)&= \frac{\mathcal{L}_{CE}(\pmb{x})}{\ln(2)\cdot n_{\text{bytes}}}
\end{align}
where the uncertainty over the data $\pmb{x}$ as measured by the sum of the cross-entropy loss is normalized by the total number of bytes in $\pmb{x}$ and a constant.

\subsection{Transformer Architecture Hyperparameters} For all the transformer blocks in \model{}, i.e. both local and global models, we largely follow the architecture of \llama{}~3~\citep{dubey2024llama}; we use the SwiGLU activation function~\citep{swiglu} in the feed-forward layers, rotary positional embeddings (RoPE)~\citep{RoPe} with $\theta=500000$~\citep{xiong2024effective} only in self-attention layers, and RMSNorm \citep{RMSNorm} for layer normalization. We use Flash attention~\citep{dao2022flashattention} for all self-attention layers that use fixed-standard attention masks such as \textit{block causal} or \textit{fixed-window block causal}, and a window size of 512 for fixed-width attention masks. Since our cross-attention layers involve dynamic patch-dependent masks, we use Flex Attention\footnote{\url{https://pytorch.org/blog/flexattention}} to produce fused implementations and significantly speed up training.

\subsection{\model{}-Specific Hyperparameters} To study the effectiveness of \model{} models, we conduct experiments along two directions, scaling trends, and downstream task evaluations, and we consider models at different scales: 400M, 1B, 2B, 4B and 8B for these experiments. The architecture hyperparameters for these models are presented in Appendix~Table~\ref{tab:arch}. 
We use max-pooling to initialize the queries for the first cross-attention layer in the local encoder. We use $500,000$ hashes with a single hash function, with n-gram sizes ranging from 3 to 8, for all  \model{} models. We use a learning rate of $4e-4$ for all models. The choice of matching learning rate between token and \model{} models follows a hyperparameter search between $1e-3$ and $1e-4$ at 400M and 1B model scales showing the same learning rate is optimal.
For scaling trends on Llama-2 data, we use training batch-sizes as recommended by~\cite{dubey2024llama} or its equivalent in bytes. For optimization, we use the AdamW optimizer~\citep{adamw} with $\beta_1$ set to 0.9 and $\beta_2$ to 0.95, with an $\epsilon=10^{-8}$. We use a linear warm-up of 2000 steps with an cosine decay schedule of the learning rate to 0, we apply a weight decay of 0.1, and global gradient clipping at a threshold of 1.0.

\section{Scaling Trends}
\label{section:scaling}

We present a holistic picture of the scaling trends of byte-level models that can inform further scaling of \model{} models.
Our scaling study aims to address the limitations of previous research on byte-level models \todoArti{include the above paragraph in the related work section and reference here} in the following ways: (a) We compare trends for the compute-optimal training regime, (b) We train matching 8B models on non-trivial amounts of training data (up to 1T tokens/4T bytes) and evaluate on downstream tasks, and (c) We measure scaling trends in inference-cost controlled settings.
In a later section, we will investigate specific advantages from modeling byte-sequences.
\begin{figure}[t]
    \centering
    \begin{subfigure}{0.5\textwidth}
        \centering
        \includegraphics[width=\linewidth]{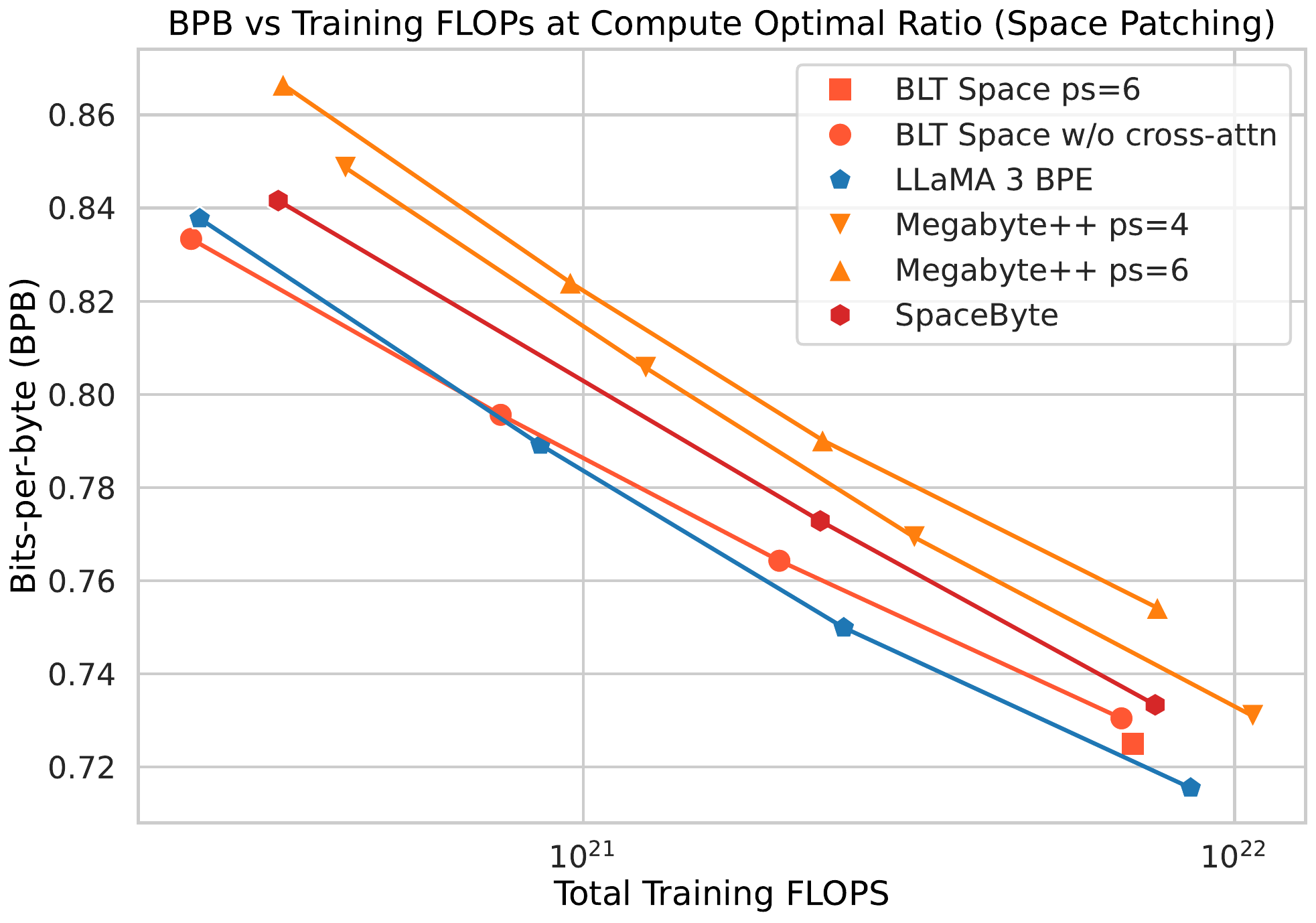}
    \end{subfigure}%
    \hfill
    \begin{subfigure}{0.5\textwidth}
        \centering
        \includegraphics[width=\linewidth]{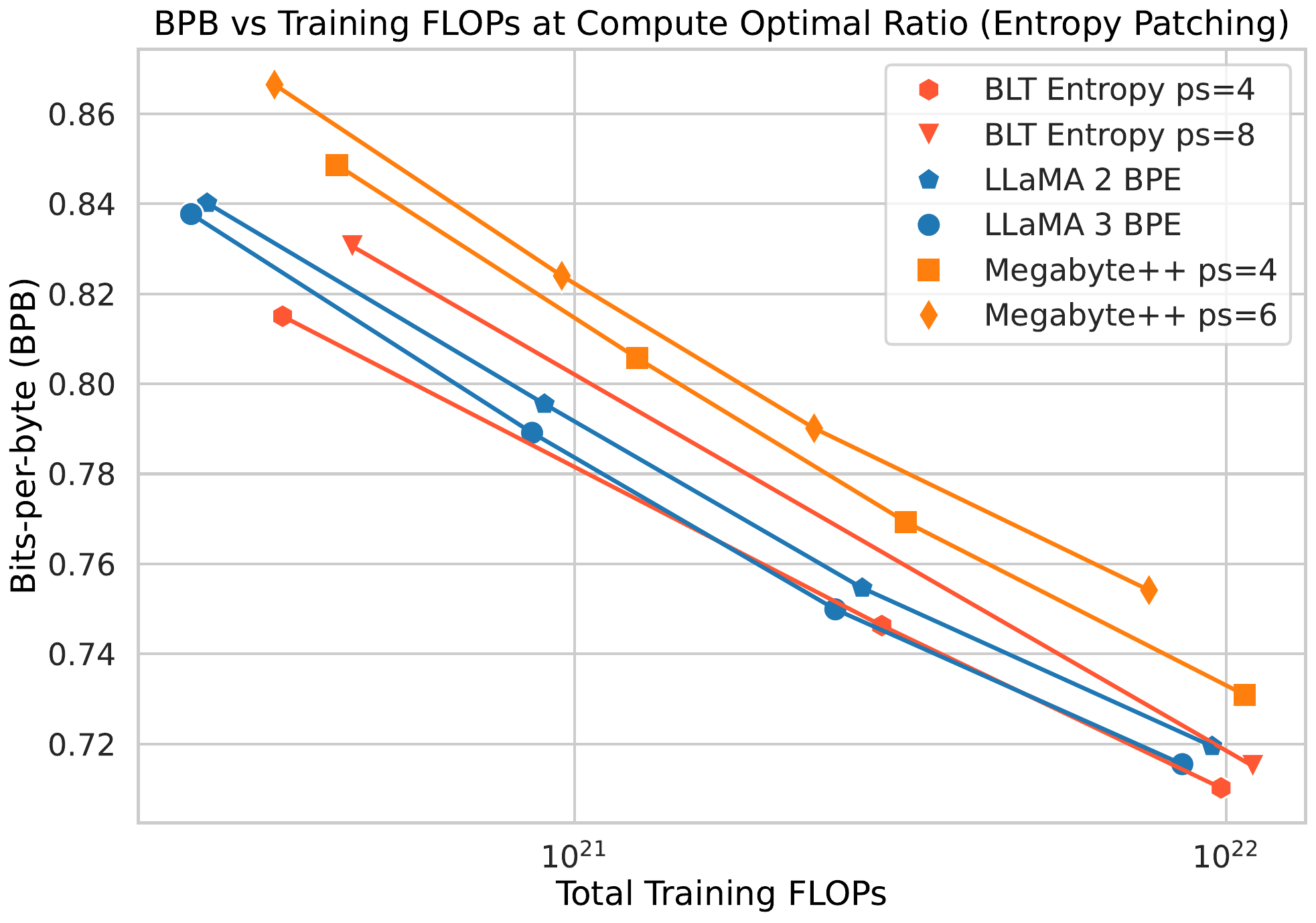}     
    \end{subfigure}
    \caption{Scaling trends for \model{} models with different architectural choices, as well as for baseline BPE token-based models. We train models at multiple scales from 1B up to 8B parameters for the optimal number of tokens as computed by~\cite{dubey2024llama} and report bits-per-byte on a sample from the training distribution. \model{} models perform on par with state-of-the-art tokenizer-based models such as \llama{} 3, at scale. PS denotes patch size. We illustrate separate architecture improvements on space-patching (\textbf{left}) and combine them with dynamic patching (\textbf{right}).}
    \label{fig:scaling-compute-opt}
\end{figure}

\subsection{Parameter Matched Compute Optimal Scaling Trends}
\label{section:parameter-matched-scaling}
Using the \llama{} 2 dataset, we train various \textit{compute-optimal} \bpe{} and \model{} models across four different sizes, ranging from 1B to 8B parameters. We then plot the training \flop{}s against language modeling performance on a representative subset of the training data mixture.
The \bpe{} models are trained using the optimal ratio of model parameters to training data, as determined by \llama{}~3~\citep{dubey2024llama}. This \textit{compute-optimal} setup is theoretically designed to achieve the best performance on the training dataset within a given training budget~\citep{hoffmann2022training}, providing a robust baseline for our model.
For each \bpe{} model, we also train a corresponding \model{} model on the same data, using a Latent Transformer that matches the size and architecture of the corresponding \bpe{} Transformer.

As illustrated in~\autoref{fig:scaling-compute-opt} (right), \model{} models either match or outperform their \bpe{} counterparts and this trend holds as we scale model size and \flop{}s. To the best of our knowledge, \model{} is the first byte-level Transformer architecture to achieve matching scaling trends with BPE-based models at compute optimal regimes. This therefore validates our assumption that the optimal ratio of parameters to training compute for \bpe{} also applies to \model{}, or at least it is not too far off.     

Both architectural improvements and dynamic patching are crucial to match \bpe{} scaling trends. In ~\autoref{fig:scaling-compute-opt} (left), we compare space-patching-based models against \llama{} 3. We approximate SpaceByte \citep{slagle2024spacebyte} using \model{} space-patching without n-gram embeddings and cross-attention. 
Although SpaceByte improves over Megabyte, it remains far from  \llama{} 3. In \autoref{fig:scaling-compute-opt} (right), we illustrate the improvements from both architectural changes and dynamic patching. \model{} models perform on par with state-of-the-art tokenizer-based models such as  \llama{} 3, at scale.
 
We also observe the effects of the choice of tokenizer on performance for tokenizer-based models, i.e., models trained with the \llama{}-3 tokenizer outperform those trained using the \llama{}-2 tokenizer on the same training data.

Finally, our \model{} architecture trends between \llama{} 2 and 3 when using significantly larger patch sizes. The \bpe{} tokenizers of \llama{} 2 and 3 have an average token size of 3.7 and 4.4 bytes. In contrast, \model{} can achieve similar scaling trends with an average patch size of 6 and even 8 bytes. Inference \flop{} are inversely proportional to the average patch size, so using a patch size of 8 bytes would lead to nearly 50\% inference \flop{} savings. 
Models with larger patch sizes also seem to perform better as we scale model and data size. \model{} with patch size of 8 starts at a significantly worse point compared to \bpe{} \llama{} 2 at 1B but ends up better than \bpe{} at 7B scale. This suggests that such patch sizes might perform better at even larger scales and possibly that even larger ones could be feasible as model size and training compute grow. 

\todoArti{this would be the right place to discuss inference time scaling.}

\subsection{Beyond Compute Optimal Task Evaluations}
\label{section:evals}
To assess scaling properties further, we train an 8B \model{} model beyond the compute optimal ratio on the \model{}-1T dataset, a larger higher-quality dataset, and measure performance on a suite of standard classification and generation benchmarks. For evaluation, we select the following common sense reasoning, world knowledge, and code generation tasks:
\paragraph{Classification tasks} include ARC-Easy (0-shot)~\citep{Eval_ARC}, Arc-Challenge (0-shot)~\citep{Eval_ARC}, HellaSwag (0-shot)~\citep{Eval_hellaswag}, PIQA (0-shot)~\citep{Eval_piqa}, and MMLU (5-shot)~\citep{hendrycks2020measuring}.  We employ a prompt-scoring method, calculating the likelihood over choice characters, and report the average accuracy.
\paragraph{Coding related generation tasks:}  We report pass@1 scores on MBPP (3-shot)~\citep{Eval_MBPP} and HumanEval (0-shot)~\citep{Eval_HumanEval}, to evaluate the ability of LLMs to generate Python code.

% \usepackage{booktabs}

\begin{table}[t]
\centering
\begin{tabular}{@{}lccc@{}}
\toprule
 & 
 \multicolumn{1}{c}{\begin{tabular}[c]{@{}c@{}}Llama 3\\ 1T Tokens\end{tabular}} & 
 \multicolumn{1}{c}{\begin{tabular}[c]{@{}c@{}}\model{}-Space \\ 6T Bytes\end{tabular}} &
 \multicolumn{1}{c}{\begin{tabular}[c]{@{}c@{}}\model{}-Entropy \\ 4.5T Bytes \end{tabular}}
 \\ \midrule
\textbf{Arc-E} &  
77.6 & 75.4 & \textbf{79.6} \\
\textbf{Arc-C} & 
\textbf{53.3} & 49.8 & 52.1  \\
\textbf{HellaSwag} & 
79.1 & 79.6 & \textbf{80.6}  \\
\textbf{PIQA} & 
80.7 & \textbf{81.1} & 80.6 \\
\textbf{MMLU} & 
\textbf{58.1} & 54.8 & 57.4 \\
\textbf{MBPP} & 
40.2 & 37.6 & \textbf{41.8} \\
\textbf{HumanEval} & 
31.1 & 27.4 & \textbf{35.4} \\ 
\midrule 
\textbf{Average} & 60.0 & 58.0 & \textbf{61.1} \\ 
\midrule
\textbf{Bytes/Patch on Train Mix} & 4.4 & \textbf{6.1} & 4.5 \\ 
\bottomrule
\end{tabular}
\caption{Comparison of \flop{}-matched \model{} \textbf{8B} models trained on the \model{}-1T dataset comprising high-quality tokens of text and code from publicly available sources, with baseline models using the \llama{} 3 tokenizer. \model{} performs better than \llama{} 3 on average, and depending on the patching scheme, achieves significant \flop{}s savings with a minor reduction in performance.}
\label{tab:evals}
\end{table}

In \autoref{tab:evals}, we compare three models trained on the \model{}-1T dataset: a \bpe{} \llama{} 3 tokenizer-based model,\footnote{We choose the \llama{} 3 tokenizer with its 128k vocabulary as it performs better than \llama{} 2's 32k vocabulary.} and two variants of the \model{} model. One employing a space-patching scheme (\model{}-Space) and another utilizing an entropy-based patching scheme (\model{}-Entropy). with approx. monotonicity constraint and reset the context of the entropy model with new lines (as discussed in~\autoref{section:entropy-context}). All three models are trained with an equivalent \flop{} budget. However, with \model{}-Entropy we additionally make an inference time adjustment of the entropy threshold from 0.6 to 0.1 which we find to improve task performance at the cost of more inference steps.

The \model{}-Entropy model outperforms the \llama{} 3 model on 4 out of 7 tasks while being trained on the same number of bytes. This improvement is like due to a combination of (1) a better use of training compute via dynamic patching, and (2) the direct modeling of byte-level information as opposed to tokens.

On the other hand, \model{}-Space underperforms the \llama{} 3 tokenizer on all but one task, but it achieves a significant reduction in inference \flop{}s with its larger average patch size of 6 bytes.
In comparison, the \bpe{} and entropy-patching based models have roughly equivalent average patch size of approximately 4.5 bytes on the training data mix. With the same training budget, the larger patch size model covers 30\% more data than the other two models which might push \model{} further away from the compute-optimal point.

\begin{table}[t]
    \centering
    \resizebox{\textwidth}{!}{
    \begin{tabular}{lllllll}
    \toprule
        \llama{} 2 & \llama{} 3 & Entropy ps=6 & Entropy ps=8 & Inference \flop{}s & Compute Optimal (Bytes) & Crossover (Bytes)\\
        \midrule
        470m  & 450m  & 610m~(1.2x)  & 760m~(1.6x)  & 3.1E8 & 50B  & 150B \\
        3.6B  &  3.9B & 5.2B~~(1.3x) & 6.6B~~(1.7x) & 2.1E9 & 400B & 1T \\
    \bottomrule
    \end{tabular}
    }
    \caption{Details of models used in the fixed-inference scaling study. We report non-embedding parameters for each model and their relative number compared to \llama{} 2. We pick model sizes with equal inference \flop{}s per byte. We also indicate BPE's compute-optimal training data quantity and the crossover point where \model{} surpasses BPE as seen in \autoref{fig:fixed_inference_scaling} (both expressed in bytes of training data). This point is achieved at much smaller scales compared to many modern training budgets.}
    \label{tab:fixed-inf-params}
\end{table}

\subsection{Patches Scale Better Than Tokens}
With \model{} models, we can simultaneously increase model size and patch size while maintaining the same training and inference \flop{} budget and keeping the amount of training data constant. Arbitrarily increasing the patch size is a unique feature of patch-based models which break free of the efficiency tradeoffs of fixed-vocabulary token-based models, as discussed in Section~\ref{section:bpe-patch}. Longer patch sizes save compute, which can be reallocated to grow the size of the global latent transformer, because it is run less often.

% \usepackage{booktabs}
\begin{table}[t]
\centering
\begin{tabular}{@{}lccc@{}}
\toprule
 &  
 \multicolumn{1}{c}{\begin{tabular}[c]{@{}c@{}}Llama 3\\ (1T tokens)\end{tabular}} &
 \multicolumn{1}{c}{\begin{tabular}[c]{@{}c@{}}Llama 3.1\\ (16T tokens)\end{tabular}} &
 \multicolumn{1}{c}{\begin{tabular}[c]{@{}c@{}}\model{} \\ (1T tokens)\end{tabular}} \\ \midrule
\textbf{HellaSwag Original} & 79.1 & \emph{80.7} & \textbf{80.6}\\
 \textbf{HellaSwag Noise Avg.} & 56.9 & 64.3 & \emph{\textbf{64.3}}\\
  \hspace{0.5cm}\textbf{- AntSpeak} &
45.6 & \emph{61.3} & \textbf{57.9} \\
\hspace{0.5cm}\textbf{- Drop} & 
53.8 & 57.3 & \emph{\textbf{58.2}} \\
\hspace{0.5cm}\textbf{- RandomCase} & 
55.3 & 65.0 & \emph{\textbf{65.7}}  \\
\hspace{0.5cm}\textbf{- Repeat} & 
57.0 & 61.5 & \emph{\textbf{66.6}}\\
\hspace{0.5cm}\textbf{- UpperCase} & 
72.9 & 76.5 & \emph{\textbf{77.3}}\\ 
\midrule
\textbf{Phonology-G2P} & 
11.8 & \emph{18.9} & \textbf{13.0} \\
\midrule
\textbf{CUTE} & 
27.5 & 20.0 & \emph{\textbf{54.1}} \\
        \hspace{0.5cm}\textbf{- Contains Char}    & 0.0	          & 0.0 & \emph{\textbf{55.9}} \\
        \hspace{0.5cm}\textbf{- Contains Word}    & 55.1	      & 21.6 & \emph{\textbf{73.5}} \\
        \hspace{0.5cm}\textbf{- Del Char}         & 34.6	      & 34.3 & \emph{\textbf{35.9}} \\
        \hspace{0.5cm}\textbf{- Del Word}         & \textbf{75.5} &	\emph{84.5} & 56.1 \\
        \hspace{0.5cm}\textbf{- Ins Char}         & 7.5	          & 0.0 & \emph{\textbf{7.6}} \\
        \hspace{0.5cm}\textbf{- Ins Word}         & \textbf{33.5} & \emph{63.3}	& 31.2 \\
        \hspace{0.5cm}\textbf{- Orthography}      & 43.1	      & 0.0 & \emph{\textbf{52.4}} \\
        \hspace{0.5cm}\textbf{- Semantic}         & 65	          & 0.0 & \emph{\textbf{90.5}} \\
        \hspace{0.5cm}\textbf{- Spelling}         & 1.1	          & - & \emph{\textbf{99.9}} \\
        \hspace{0.5cm}\textbf{- Spelling Inverse} & 30.1	      & 3.6 & \emph{\textbf{99.9}} \\
        \hspace{0.5cm}\textbf{- Substitute Char}  & 0.4	          & 1.2 & \emph{\textbf{48.7}} \\
        \hspace{0.5cm}\textbf{- Substitute Word}  & 16.4	      & 6.8 & \emph{\textbf{72.8}} \\
        \hspace{0.5cm}\textbf{- Swap Char}        & 2.6	          & 2.4 & \emph{\textbf{11.5}} \\
        \hspace{0.5cm}\textbf{- Swap Word}        & 20.1	      & 4.1 & \emph{\textbf{21}} \\

\bottomrule
\end{tabular}
\caption{We compare our 8B \model{} model to 8B BPE \llama{} 3 trained on 1T tokens on tasks that assess robustness to noise and awareness of the constituents of language  (best result bold). We also report the performance of \llama{} 3.1 on the same tasks and underline best result overall. \model{} outperforms the \llama{} 3 BPE model by a large margin and even improves over \llama{} 3.1 in many tasks indicating that the byte-level awareness is not something that can easily be obtained with more data.}
\label{tab:char_tasks}
\end{table}

We conduct a fixed inference scaling study to test the hypothesis that larger models taking fewer steps on larger patches might perform better than smaller models taking more steps. Starting from model sizes of 400m and 3.6B parameters with the \llama{} 2 tokenizer, we find \flop{} equivalent models with the \llama{} 3 tokenizer and \model{}-Entropy models with average patch sizes of 6 and 8 bytes on the training datamix (see~\autoref{tab:fixed-inf-params} for model details). For patch size 8 models, we use 3 encoder layers instead of 1. We train each model for various training \flop{} budgets.

\autoref{fig:fixed_inference_scaling} shows that \model{} models achieve better scaling trends than tokenization-based architectures for both inference \flop{} classes. In both cases, BPE models  perform better with small training budgets and are quickly surpassed by \model{}, not far beyond the compute-optimal regime. In practice, it can be preferable to spend more during the one-time pretraining to achieve a better performing model with a fixed inference budget. A perfect example of this is the class of 8B models, like \llama{} 3.1, which has been trained on two orders of magnitude more data than what is compute-optimal for that model size.

The crossover point where \model{} improves over token-based models has shifted slightly closer to the compute-optimal point when moving to the larger \flop{} class models (from 3x down to 2.5x the compute optimal budget). Similarly, the larger patch size 8 model has steeper scaling trend in the larger \flop{} class overtaking the other models sooner.
As discussed in Section~\ref{section:parameter-matched-scaling}, larger patch sizes appear to perform closer to BPE models at larger model scales. We attribute this, in part, to the decreasing share of total \flop{}s used by the byte-level Encoder and Decoder modules which seem to scale slower than the Latent Transformer. When growing total parameters 20x from 400M to 8B, we only roughly double \model{}'s local model parameters. This is important as larger patch sizes only affect \flop{}s from the patch Latent Transformer and not the byte-level modules. In fact, that is why the \model{}-Entropy ps=8 went from 1.6x to 1.7x of the \llama{} 2 model size when moving to the larger model scale.

In summary, our patch-length scaling study demonstrates that the \model{} patch-based architecture can achieve better scaling trends by simultaneously increasing both patch and model size. Such trends seem to persist and even improve at larger model scales.
\todoArti{We have two experiments running on the \model{}-T dataset with the same inference \flop{}s as BPE model. They should be done on Thursday mid-day.}

\begin{table}[t]
\centering
\begin{tabular}{@{}lcccc@{}}
\toprule

 Language & \multicolumn{2}{c}{Language $\rightarrow$ English} & \multicolumn{2}{c}{English $\rightarrow$ Language} \\ 
\midrule
& \llama{} 3 & \model{} & \llama{} 3 & \model{} \\
 \midrule
 \textbf{Arabic} & 22.3 & 24.6 & 10.4  & 8.8 \\
 \textbf{German} & 41.3 & 42.0 & 29.8  & 31.2  \\
 \textbf{Hindi} & 20.7 & 20.9 & 7.8  & 7.2 \\ 
 \textbf{Italian} &  34.0  & 33.9 & 24.4 & 26.2 \\ 
 \textbf{Vietnamese} & 31.2  & 31.0 & 28.4 & 23.7 \\
 \textbf{Thai} &  17.9  & 18.1 & 10.5 & 7.7 \\
 
\midrule

\textbf{Armenian} & 1.7 & 6.3 & 0.6 & 0.9 \\
\textbf{Amharic} & 1.3  & 3.1 & 0.4 & 0.5 \\
\textbf{Assamese} & 2.7  & 5.4 & 0.8 & 1.6 \\
\textbf{Bengali} & 4.7 & 12.7 & 1.7 & 4.1 \\
\textbf{Bosnian} & 36.0 & 37.3 & 16.9 & 19.6 \\
\textbf{Cebuano} & 18.2 & 20.6 & 5.8 & 9.1 \\
\textbf{Georgian} & 1.7 & 7.4 & 1.0 & 2.5 \\
\textbf{Gujarati} & 2.0 & 5.8 & 1.0 & 2.2 \\
\textbf{Hausa} & 5.75 & 5.9 & 1.2 & 1.3 \\
\textbf{Icelandic} & 16.1 & 17.9 & 4.8 & 5.3 \\
\textbf{Kannada} & 1.6 & 3.9 & 0.7 &  1.7 \\
\textbf{Kazakh} & 5.6 & 7.0 & 1.0 & 2.6 \\
\textbf{Kabuverdianu} & 20.3 & 20.9 & 5.1 & 6.8 \\
\textbf{Khmer} & 4.4 & 9.5 & 0.8 & 0.8 \\
\textbf{Kyrgyz} & 4.6 & 5.1 & 0.9 & 2.0 \\
\textbf{Malayalam} & 1.8 & 3.5 & 0.7 & 1.4 \\
\textbf{Odia} & 1.6 & 2.7 & 0.8 & 1.1 \\
\textbf{Somali} & 5.0  & 5.0 & 1.1 & 1.4 \\
\textbf{Swahili} & 10.1 & 12.0 & 1.4 & 2.3 \\
\textbf{Urdu} & 9.3 & 9.5 & 2.0 & 1.4 \\
\textbf{Zulu} & 4.7 & 5.0 & 0.6 & 0.5 \\
\midrule
\textbf{Overall Average} & 12.1 & \textbf{14.0} & 5.9 & \textbf{6.4} \\
\bottomrule
\end{tabular}
\caption{Performance of 8B \model{} and 8B Llama 3 trained for 1T tokens on translating into and from six widely-used languages and twenty one lower resource languages with various scripts from the FLORES-101 benchmark \citep{goyal2022flores}. }
\label{tab:flores}
\end{table}

\section{Byte Modeling Improves Robustness}
We also measure the robustness of \model{} compared to token-based models that lack direct byte-level information, and present an approach to byte-ify pretrained token-based models.
\label{sec:robustness}

\subsection{Character-Level Tasks}

A very early motivation for training byte-level models was to take advantage of their robustness to byte level noise in the input, and also to exploit their awareness of the constituents of tokens, which current tokenizer-based models struggle with. To measure these phenomena, we perform additional evaluations on benchmarks that evaluate both robustness to input noise as well as awareness of characters, both English and multi-lingual, including digits and phonemes. We present these results in Table~\ref{tab:char_tasks}.

\paragraph{Noisy Data} We create noised versions of the benchmark classification tasks described in Section~\ref{section:evals}, to compare the robustness of tokenizer-based models with that of \model{}.
We employ five distinct character-level noising strategies to introduce variations in the text: (a) \textit{AntSpeak}: This strategy converts the entire text into uppercase, space-separated characters. (b) \textit{Drop}: Randomly removes 10\% of the characters from the text. (c) \textit{RandomCase}: Converts 50\% of the characters to uppercase and 50\% to lowercase randomly throughout the text. (d) \textit{Repeat}:  Repeats 20\% of the characters up to a maximum of four times. (e) \textit{UpperCase}: Transforms all characters in the text to uppercase.
During evaluation, we apply each noising strategy to either the prompt, completion, or both as separate tasks and report the average scores. In Table \ref{tab:char_tasks} we report results on noised HellaSwag~\citep{Eval_hellaswag} and find that \model{} indeed outperforms tokenizer-based models across the board in terms of robustness, with an average advantage of 8 points over the model trained on the same data, and even improves over the \llama{} 3.1 model trained on a much larger dataset.

\paragraph{Phonology - Grapheme-to-Phoneme (G2P)}
We assess \model{}'s capability to map a sequence of graphemes (characters representing a word) into a transcription of that word's pronunciation (phonemes). In Table \ref{tab:char_tasks}, we present the results of the G2P task in a 5-shot setting using Phonology Bench~\citep{suvarna2024phonologybench} and find that \model{} outperforms the baseline \llama{} 3 1T tokenizer-based model on this task.

\begin{figure}
    \centering
    \resizebox{\columnwidth}{!}{%
        \begin{tabular}{p{0.12\textwidth}|p{0.5\textwidth}|p{0.15\textwidth}|p{0.15\textwidth}}
            \toprule
             \textbf{Task} & \textbf{Prompt} & \textbf{\llama{}} 3 &  \textbf{\model{}} \\
             \midrule
             \texttt{Substitute Word} & \texttt{Question: Substitute " and " with " internet " in " She went to the kitchen and saw two cereals. ". Answer:} & \texttt{She went to the kitchen and saw two cereals.} & \texttt{She went to the kitchen internet saw two cereals.} \\
             \hline
             \texttt{Swap Char} & \texttt{Question: Swap " h " and " a " in " that ". Answer: } & \texttt{that} & \texttt{taht} \\ 
             \hline
             \texttt{Substitute Char} & \texttt{Question: Substitute " a " with " m " in " page ". Answer: } & \texttt{-} & \texttt{pmge} \\
             \hline
             \texttt{Semantic Similarity} & \texttt{Question: More semantically related to " are ": " seem ", " acre ". Answer: } & \texttt{acre} & \texttt{seem} \\
             \hline
             \texttt{Orthographic Similarity} & \texttt{Question: Closer in Levenshtein distance to " time ": " timber ", " period ". Answer: } & \texttt{period} & \texttt{timber} \\
             \hline
             \texttt{Insert Char} & \texttt{Question: Add an " z " after every " n " in " not ". Answer: } & \texttt{znotz} & \texttt{nzot} \\
             \bottomrule
        \end{tabular}
        }
        \caption{Output responses from \llama{} 3 and \model{} models for various tasks from CUTE benchmark. \model{} model performs better on sequence manipulation tasks compared to the tokenizer-based \llama{} 3 model. Note that few-shot examples are not shown in the above prompts to maintain clarity.}
    \label{fig:cute_examples}
\end{figure}

\paragraph{CUTE}
To assess character-level understanding, we evaluate \model{} on the CUTE benchmark~\citep{edman2024cute}, which comprises several tasks that are broadly classified into three categories: understanding composition, understanding orthographic similarity, and ability to manipulate sequences. This benchmark poses a significant challenge for most tokenizer-based models, as they appear to possess knowledge of their tokens' spellings but struggle to effectively utilize this information to manipulate text. Table~\ref{tab:char_tasks} shows that \model{}-Entropy outperforms both BPE \llama{} 3 models by more than 25 points on this benchmark. In particular, our model demonstrates exceptional proficiency in character manipulation tasks achieving 99.9\% on both spelling tasks. Such large improvements despite \model{} having been trained on 16x less data than \llama{} 3.1 indicates that character level information is hard to learn for BPE models.
Figure \ref{fig:cute_examples} illustrates a few such scenarios where \llama{} 3 tokenizer model struggles but our \model{} model performs well. 
Word deletion and insertion are the only two tasks where BPE performs better. Such word manipulation might not be straightforward for a byte-level model but the gap is not too wide and building from characters to words could be easier than the other way around. We use the same evaluation setup in all tasks and the original prompts from Huggingface. BPE models might benefit from additional prompt engineering.

\paragraph{Low Resource Machine Translation} We evaluate \model{} on translating into and out of six popular language families and twenty one lower resource languages with various scripts from the FLORES-101 benchmark~\citep{goyal2022flores} and report SentencePiece BLEU in Table~\ref{tab:flores}.
Our results demonstrate that \model{} outperforms a model trained with the \llama{} 3 tokenizer, achieving a 2-point overall advantage in translating into English and a 0.5-point advantage in translating from English. In popular language pairs, \model{} performs comparably to or slightly better than \llama{} 3. However, \model{} outperforms \llama{} 3 on numerous language pairs within lower-resource language families, underscoring the effectiveness of byte modeling for generalizing to long-tail byte sequences.
\todoArti{We should add a discussion on why Lang -> Engl works better. I think it's because in the other direction, we are using fewer inference flops compared to BPE - we can mention average patch size on wikipedia which is more multilingual}

% \usepackage{booktabs}
\begin{table}[t]
\centering
\begin{tabular}{@{}lcccc@{}}
\toprule
 & \multicolumn{1}{c}{\begin{tabular}[c]{@{}c@{}}Llama 3\\ 8B \\ (220B tokens)\end{tabular}} & \multicolumn{1}{c}{\begin{tabular}[c]{@{}c@{}}\model{}\\ 8B\\ (220B tokens)\end{tabular}} & \multicolumn{1}{c}{\begin{tabular}[c]{@{}c@{}}\model{}  from \llama{} 3.1 \\ 8B\\ (220B tokens)\end{tabular}} & \multicolumn{1}{c}{\begin{tabular}[c]{@{}c@{}} Llama 3.1 \\ 8B\\ (15T tokens)\end{tabular}} \\ \midrule
\textbf{Arc-E} & 67.4 & 66.8 & 66.6 & 83.4 \\
\textbf{Arc-C} & 40.4 & 38.8 & 45.8 & 55.2 \\
\textbf{HellaSwag} & 71.2 & 72.2 & 76.1 & 80.7 \\
\textbf{PIQA} & 77.0 & 78.2 & 77.4 & 80.7 \\ 
\textbf{MMLU} & 26.5 & 25.2 & 63.7 & 66.3\\ 
\textbf{MBPP} & 11.8 & 10.0 & 38.2 & 47.2 \\ 
\textbf{HumanEval} & 9.2 & 7.3 & 34.2 & 37.2 \\ 
\bottomrule
\end{tabular}
\caption{Initializing the global transformer model of \model{} from the non-embedding parameters of Llama 3 improves performance on several benchmark tasks. First three models trained on the \llama{} 2 data for compute-optimal steps.}
\label{tab:distill}
\end{table}

\subsection{Training \model{} from \llama{} 3}
\label{sec:distillation}
We explore a workflow where \model{} models can leverage existing pre-trained tokenizer-based models for better and faster training convergence, acheived by initializing the global transformer parameters of \model{} with those of a pre-trained \llama{} 3.1 model. Subsequently, we update the weights of the global transformer using one-tenth the learning rate employed for the local encoder and local decoder model, for \llama{} 3 optimal number of steps, and present a comparison with a baseline \model{} in Table~\ref{tab:distill}. It is evident that \model{} from \llama{} 3.1 significantly outperforms both the \llama{} 3 and \model{} baselines, which were trained with the same number of \flop{}s. Moreover, when compared to our \model{}-Entropy model (as presented in Table~\ref{tab:evals}), which was trained on a significantly larger dataset (1T tokens), \model{} from \llama{} 3.1 still achieves superior performance on MMLU task, suggesting that it can be an effective approach in significantly reducing the training \flop{}s. 

This setup can also be viewed as transforming tokenizer-based models into tokenizer-free ones, effectively converting a pre-trained LLaMA 3.1 model into a \model{} model. To provide a comprehensive comparison, we include the original LLaMA 3.1 model trained on 15T tokens in Table \ref{tab:distill} and evaluate it against the \model{} derived from LLaMA 3. Our model experiences a slight performance decline on MMLU and HumanEval, but a more significant drop on other tasks. This suggests that further work is needed to fully leverage the pre-trained model and improve upon its performance, particularly in terms of optimizing data mixtures and other hyperparameters.

\section{Ablations and Discussion}
\label{section:ablations}
In this section, we discuss ablations justifying architectural choices for \model{} and the patching scheme and hyper-parameters for the \model{} 8B parameter model trained on the \model{}-1T dataset.

\begin{figure}[t]
    \centering
    \includegraphics[width=0.6\textwidth]{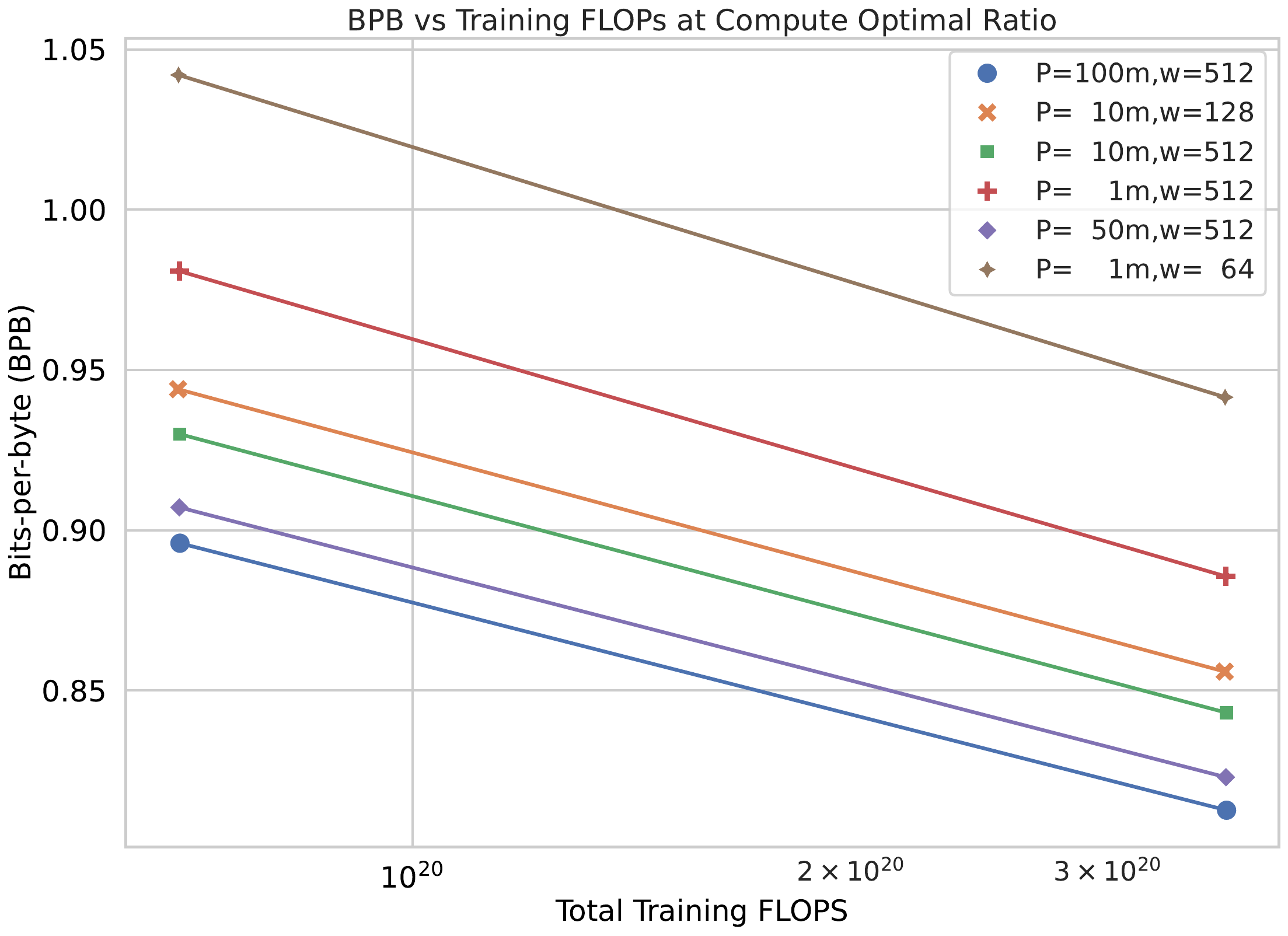}
    \caption{Variation of language modeling performance in bits-per-byte (bpb) with training \flop{}s for 400m and 1b \model{} models patched with entropy models of different sizes and context windows. Both dimensions improve scaling performance, with diminishing returns beyond 50m parameter entropy models with a context of 512 bytes.}
    \label{fig:entropy_ablation}
\end{figure}

\paragraph{Entropy Model Hyper-parameters} To study the effect of varying entropy model size and context window length on scaling performance, we train byte-level entropy transformer models of different model sizes between 1m and 100m parameters, with varying context window lengths from 64 to 512. We plot bpb vs training \flop{} scaling law curves, created using our $400m$ and $1b$ \model{} models trained on the Llama-2 dataset and present them in \autoref{fig:entropy_ablation}. We find that scaling performance is positively correlated with both these dimensions of the entropy model, with diminishing returns when we scale beyond 50m parameters. 

\paragraph{Types of Patching}
\todoArti{Add discussion about cnn and translation to lookup table also this section needs polishing}
We ablate the four different patching schemes, introduced in Section~\ref{section:patching} i.e. 1) Strided Patching with a stride of 4 and 6, 2) Patching on whitepsace, 3) BPE Tokenizer patching based on the \llama{} 3 tokenizer, and 4) Entropy based patching using a small byte \llm{}. 

% \usepackage{booktabs}
\begin{table}
\centering
\begin{tabular}{@{}lccc@{}}
\toprule
 & \multicolumn{1}{c}{\begin{tabular}[c]{@{}c@{}}Llama 3\\ BPE\end{tabular}} & \multicolumn{1}{c}{\begin{tabular}[c]{@{}c@{}}Space Patching\\ \model{} \end{tabular}} & \multicolumn{1}{l}{\begin{tabular}[c]{@{}c@{}}Entropy\\ \model{}\end{tabular}} \\ \midrule
\textbf{Arc-E} & 67.4 & 67.2 & 68.9 \\
\textbf{Arc-C} & 40.5 & 37.6 & 38.3 \\
\textbf{HellaSwag} & 71.3 & 70.8 & 72.7 \\
\textbf{PIQA} & 77.0 & 76.5 & 77.6 \\
\bottomrule
\end{tabular}
\caption{Benchmark evaluations of two patching schemes for 8b \model{} models and BPE \llama{3} baseline. These models are trained on the \llama{} 2 data for the optimal number of steps as determined by \cite{dubey2024llama}.
}
\label{tab:evals_ablation}
\end{table}

While dynamic patching reduces the effective length of sequences, we control for the sequence length to maintain a similar context length for all patching schemes. All the models see the same number of bytes in each sequence during training and inference in expectation to prevent any confounding factors from being able to model larger contexts. Figure~\ref{fig:scaling-compute-opt} highlights the results of these ablations. All the remaining patching schemes outperform static patching, with space patching being a very close competitor to dynamic entropy based patching.

In \autoref{tab:evals_ablation}, we present benchmark evaluations for \model{} models comparing tokenizer-based models, space patching, and entropy-based patching, trained on the Llama 2 dataset for an optimal number of steps~\citep{dubey2024llama}. Although space patching is a simpler strategy that does not involve running an entropy model on the fly during training, we find that the gains we observed using entropy-based patching on scaling trends~(Section~\ref{section:scaling}) do indeed carry forward even to downstream benchmark tasks.\footnote{Space patching results are from earlier runs without cross-attention, but similar trends are observed even with cross-attention.} 

\begin{table}[t]
\centering
\begin{tabular}{@{}cccrrrr@{}}
\toprule
 &  & &  \multicolumn{4}{c}{BPB}  \\
 \cmidrule(lr){4-7} 
Cross Attn. Dec. & Cross Attn. Enc. & Pooling Init &  Wikipedia &    CC &  Github &  Train Dist \\
\midrule
               - &       All Layers &        False &      0.830 & 0.915 &   \textbf{0.442} &       0.891 \\
               - &       Last Layer &        False &      0.836 & 0.906 &   0.447 &       0.886 \\
               - &                - &            - &      0.833 & 0.892 &   0.446 &       0.866 \\
     First Layer &       Last Layer &         True &      0.825 & 0.883 &   0.443 &       0.861 \\
      All Layers &       Last Layer &         True &      \textbf{0.823} & 0.871 &   0.443 &       0.846 \\
      All Layers &       All Layers &         True &      0.828 & \textbf{0.868} &   0.443 &       \textbf{0.844} \\
\bottomrule
\end{tabular}
\caption{Ablations on the use of Cross Attention for a 1B \model{} model trained on 100B bytes. We report bits-per-byte (bpb) on different datasets. We also report bpb on a random sample of the training data (denoted as Train Dist.) The Cross Attn. Enc. and Dec. columns denote which transformer layers the cross-attention block is applied after (or before for the decoder) in the local encoder and decoder respectively.
}
\label{tab:crossattn_ablations_1b}
\end{table}

\begin{table}[th]
\centering
\begin{tabular}{@{}lllrrrr@{}}
\toprule
 & & &  \multicolumn{4}{c}{BPB} \\
\cmidrule(lr){4-7} 
Ngram Sizes & Per Ngram Vocab & Total Vocab &  Wikipedia &    CC &  Github &  Train Dist \\
\midrule
               - &       - &        - &    0.892 & 0.867 &   0.506 &       0.850 \\
           6,7,8 &      100k &   300k &    0.873 & 0.860 &   0.499 &       0.842 \\
           6,7,8 &      200k &   600k &    0.862 & 0.856 &   0.492 &       0.838 \\
           3,4,5 &      100k &   300k &    0.859 & 0.855 &   0.491 &       0.837 \\
           6,7,8 &      400k &     1M &    0.855 & 0.853 &   0.491 &       0.834 \\
           3,4,5 &      200k &   600k &    0.850 & 0.852 &   0.485 &       0.833 \\
     3,4,5,6,7,8 &      100k &   600k &    0.850 & 0.852 &   0.486 &       0.833 \\
           3,4,5 &      400k &     1M &    0.844 & 0.851 &   0.483 &       0.832 \\
     3,4,5,6,7,8 &      200k &     1M &    0.840 & 0.849 &   0.481 &       0.830 \\
     3,4,5,6,7,8 &      400k &     2M &    \textbf{0.831} & \textbf{0.846} &   \textbf{0.478} &       \textbf{0.826} \\
\bottomrule
\end{tabular}
\caption{Ablations on the use of n-gram hash embedding tables for a 1B \model{} model trained on 100B bytes. We find that hash n-gram embeddings are very effective with very large improvements in BPB.
The most significant parameter is the per-ngram vocab size and that smaller ngram sizes are more impactful than larger ones.
}
\label{tab:ngram_ablations_1b}
\end{table}

\paragraph{Cross-Attention} 
In~\autoref{tab:crossattn_ablations_1b}, we ablate including cross-attention at various points in the encoder and decoder of \model{}. 
For the encoder cross-attention we test initializing the queries with 1) the same learned embedding for every global state, 2) a hash embedding of the bytes in the patch, and 3) pooling of the encoder hidden representation of the patch bytes at the given encoder layer.

We find that using cross-attention in the \textit{decoder} is most effective. In the encoder, there is a slight improvement in using cross-attention but only with pooling initialization of queries. Additionally, we find that cross-attention helps particularly on Common-Crawl and especially with larger patch sizes.

\paragraph{n-gram Hash Embeddings} 

We ablate settings of 0, 100K, 200K and 400K n-gram hash embedding vocabularies and present results in Table~\ref{tab:ngram_ablations_1b}. We find that hash embeddings help on all domains, but particularly on Wikipedia and Github (0.04 bpb difference compared to 0.01 bpb difference after 15k steps at 8B).
At 8B scale going from 500K to 300K hashes changed performance by ~0.001 bpb on 15k steps. This indicates that hashes are vital to bringing the performance of \model{} to match those of tokenizer based models, however, after 300K hashes, there are diminishing returns. Additionally, it appears that the gains are largely complementary with cross-attention as they provide improvements on different datasets.

\begin{table}[t]
\centering

\begin{tabular}{@{}lllr@{}}
\toprule

Ngram Embeddings & Encoder Layers & Decoder Layers & Train Dist BPB \\

\midrule
False &      1 &        9 &       0.850 \\
False &      5 &        5 &       0.843 \\
\midrule
True  &      5 &        5 &       0.844 \\
True  &      3 &        7 &       0.824 \\
True  &      1 &        9 &       0.822 \\
\bottomrule
\end{tabular}

\caption{
When paired with hash n-gram embeddings, a light-weight local encoder is sufficient. More layers can then be allocated to the decoder for the same cost.
}
\label{tab:local_layers}
\end{table}

\paragraph{Local Model Hyperparamaters} 
In Table \ref{tab:local_layers}, we ablate various settings for the number of layers in the local encoder and decoder. When paired with hash n-gram embeddings, \model{} works well with an encoder that is extremely light-weight i.e. just one layer, and with a heavier decoder.

\section{Related Work}
\label{section:related_work}
\todoArti{Could also relate to conditional compute literature}

\textbf{Character-Level RNNs:} Character Language Modeling has been a popular task ever since the early days of neural models~\citep{sutskever2011generating,mikolov2012subword,graves2013generating} owing to their flexibility of modeling out of vocabulary words organically without resorting to back-off methods. \cite{kim2016character} also train a model that processes characters only on the input side using convolutional and highway networks that feed into LSTM-based RNNs and are able to match performance with the RNN based state-of-the-art language models of the time on English and outperform them on morphologically rich languages, another sought-after advantage of character-level LLMs. \cite{kenter2018byte} do machine comprehension using byte-level LSTM models that outperformed word-level models again on morphologically-rich Turkish and Russian languages. Along similar lines, \cite{zhang2015character} used character-based convolutional models for classification tasks, which outperformed word-level models for certain tasks. \citet{chung2019hierarchical} use hierarchical LSTM models using boundary-detectors at each level to discover the latent hierarchy in text, to further improve performance on character level language modeling. ByteNet by \cite{kalchbrenner2016neural} uses CNN based layers on characters as opposed to attention for machine translation. 

\textbf{Character-Level Transformers:} The development of transformer models using attention~\citep{vaswani2017attention} together with subword tokenization~\citep{sennrich-etal-2016-neural}, significantly improved the performance of neural models on language modeling and benchmark tasks. However, word and sub-word units implicitly define an inductive bias for the level of abstraction models should operate on. To combine the successes of transformer models with the initial promising results on character language modeling, \citet{al2019character} use very deep transformers, and with the help of auxiliary losses, train transformer-based models that outperformed previous LSTM based character \llms{}. However, they still saw a significant gap from word level LLMs. GPT-2~\citep{radfordgpt2} also observed that on large scale datasets like the 1 billion word benchmark, byte-level LMs were not competitive with word-level LMs. 

While \cite{Choe2019BridgingTG} demonstrated that byte-level \llms{} based on transformers can outperform subword level LLMs with comparable parameters, the models take up much more compute and take much longer to train. Similarly, \cite{el2020characterbert} train a BERT model (CharFormer) that builds word representations by applying convolutions on character embeddings, and demonstrate improvements on the medical domain, but they also expend much more compute in doing so. \cite{clark2022canine} develop CANINE, a 150M parameter encoder-only model that operates directly on character sequences. CANINE uses a deep transformer stack at its core similar in spirit to our global model, and a combination of a local transformer and strided convolutions to downsample the input characters, and outperforms the equivalent token-level encoder-only model (mBERT) on downstream multilingual tasks. ByT5~\citep{xue2022byt5} explored approaches for byte-level encoder decoder models, that do not use any kind of patching operations. While their model exhibited improved robustness to noise, and was competitive with tokenizer-based models with 4x less data, the lack of patching meant that the models needed to compute expensive attention operations over every byte, which was extremely compute heavy. Directly modeling bytes instead of subword units increases the sequence length of the input making it challenging to efficiently scale byte level models. Recently, using the Mamba Architecture~\citep{gu2023mamba},  which can maintain a fixed-size memory state over a very large context length, \cite{wang2024mambabyte} train a byte-level Mamba architecture also without using patching, and are able to outperform byte-level transformer models in a \flop{} controlled setting at the 350M parameter scale in terms of bits-per-byte on several datasets.

\textbf{Patching-based approaches:} The effective use of patching can bring down the otherwise inflated number of \flop{}s expended by byte-level LLMs while potentially retaining performance, and many works demonstrated initial successes at a small scale of model size and number of training bytes. \cite{nawrot-etal-2022-hierarchical} experiment with static patching based downsampling and upsampling and develop the hourglass transformer which outperforms other byte-level baselines at the 150M scale. \cite{nawrot-etal-2023-efficient} further improve this with the help of dynamic patching schemes, including a boundary-predictor that is learned in an end-to-end fashion, a boundary-predictor supervised using certain tokenizers, as well as an entropy-based patching model similar to \model{}, and show that this approach can outperform the vanilla transformers of the time on language modeling tasks at a 40M parameter scale on ~400M tokens. \cite{lester2024training} investigate training on sequences compressed using arithmetic coding to achieve compression rates beyond what BPE can achieve, and by using a equal-info windows technique, are able to outperform byte-level baselines on language modeling tasks, but underperform subword baselines.
\todoArti{Add MyT5 MrT5 Magnet languages cost the same}
% https://arxiv.org/abs/2407.08818
% https://arxiv.org/abs/2305.13707
% https://arxiv.org/abs/2410.20771

Our work draws inspiration and is most closely related to MegaByte~\citep{yu2023megabyte}, which is a decoder only causal LLM that uses a fixed static patching and concatenation of representations to convert bytes to patches, and uses a local model on the decoder side to convert from patches back into bytes. They demonstrate that MegaByte can match tokenizer-based models at a 1B parameter scale on a dataset of ~400B bytes. We ablate MegaByte in all our experiments and find that static patching lags behind the current state-of-the-art compute optimally trained tokenizer based models in a \flop{} controlled setting and we demonstrate how \model{} bridges this gap. \cite{slagle2024spacebyte} make the same observation about MegaByte and suggest extending the static patching method to patching on whitespaces and other space-like bytes, and also add a local encoder model. They find improvements over tokenized-based transformer models in a compute controlled setting on some domains such as Github and arXiv at the 1B parameter scale. We also report experiments with this model, and show that further architectural improvements are needed to scale up byte-level models even further and truly match current state-of-the-art token-based models such as \llama{} 3.

\section{Limitations and Future Work}

In this work, for the purposes of architectural choices, we train models for the optimal number of steps as determined for \llama{}~3~\citep{dubey2024llama}. However, these scaling laws were calculated for BPE-level transformers and may lead to suboptimal (data, parameter sizes) ratios in the case of \model{}. We leave for future work the calculation of scaling laws for \model{} potentially leading to even more favorable scaling trends for our architecture. Additionally, many of these experiments were conducted at scales upto 1B parameters, and it is possible for the optimal architectural choices to change as we scale to 8B parameters and beyond, which may unlock improved performance for larger scales. 

Existing transformer libraries and codebases are designed to be highly efficient for tokenizer-based transformer architectures. While we present theoretical \flop{} matched experiments and also use certain efficient implementations (such as FlexAttention) to handle layers that deviate from the vanilla transformer architecture, our implementations may yet not be at parity with tokenizer-based models in terms of wall-clock time and may benefit from further optimizations. 

While \model{} uses a separately trained entropy model for patching, learning the patching model in an end-to-end fashion can be an interesting direction for future work. In Section \ref{sec:distillation}, we present initial experiments showing indications of success for ``byte-ifying'' tokenizer-based models such as Llama 3 that are trained on more than 10T tokens, by initializing and freezing the global transformer with their weights. Further work in this direction may uncover methods that not only retain the benefits of bytefying, but also push performance beyond that of these tokenizer-based models without training them from scratch.

\section{Conclusion}
\label{section:conclusion}

This paper presents the Byte Latent Transformer (\modelbf{}), a new architecture that redefines the conventional dependency on fixed-vocabulary tokenization in large language models. By introducing a dynamic, learnable method for grouping bytes into patches, \model{} effectively allocates computational resources based on data complexity, leading to significant improvements in both efficiency and robustness. Our extensive scaling study demonstrates that \model{} models can match the performance of tokenization-based models  like \llama{} 3 at scales up to 8B and 4T bytes, and can trade minor losses in evaluation metrics for up to 50\% reductions in inference \flop{}s. 
Furthermore, \model{} unlocks a new dimension for scaling, allowing simultaneous increases in model and patch size within a fixed inference budget. This new paradigm becomes advantageous for compute regimes commonly encountered in practical settings. 
While directly engaging with raw byte data, \model{} also improves the model's ability to handle the long-tail of data, offering significant improvements in robustness to noisy inputs and a deeper understanding of sub-word structures. 
Overall, these results position \model{} as a promising alternative to traditional tokenization-based approaches, providing a scalable and robust framework for more efficient and adaptable language models.

\section*{Acknowledgements}

We would like to thank Kalyan Saladi for help with everything relating to pre-training infrastructure; Gabriel Synnaeve, Ammar Rizvi, Jacob Kahn, Michel Meyer for helping organize resources for scaling up \model{}; Badr Youbi Idirissi, Mathurin Videau, and Jade Copet for invaluable discussions and feedback about \model{}, for access to the Lingua framework for open-sourcing code for \model{}, and for help preparing the \model{}-1T dataset used in this paper; Omer Levy, who was actively involved in the early stages of the project and provided valuable feedback and ideas; Driss Guessous for help with FlexAttention; and Sida Wang, Melanie Sclar, Amanda Bertsch, and Hunter Lang for feedback and discussions.

\section*{Contributors}
\definecolor{starcolor}{RGB}{255,127,80} % coral for joint first authors
\definecolor{daggercolor}{RGB}{25,25,112} % midnight blue for key contributors
\definecolor{ddaggercolor}{RGB}{0,100,0} % dark green for workstream leads
\definecolor{sharpcolor}{RGB}{160,0,0} % dark red for project leads

In this section, we list individual contributions.

\paragraph{\textbf{Core Contributors:}} Artidoro Pagnoni, Srinivasan Iyer, Ramakanth Pasunuru, Pedro Rodriguez, John Nguyen, Gargi Ghosh (Project Lead)

\vspace{0.2em}

\paragraph{\textbf{Core Advising Group:}} Mike Lewis, Ari Holtzman, Luke Zettlemoyer

\vspace{0.2em}

\paragraph{\textbf{Advisors and Contributors:}} Jason Weston, Benjamin Muller, Margaret Li, Chunting Zhou, Lili Yu

\clearpage
\newpage
\bibliographystyle{assets/plainnat}
\bibliography{journal-full,paper}

\clearpage
\newpage
\beginappendix
\label{section:appendix}

\section{Model Hyper Parameters}

Table~\ref{tab:arch} shows different hyper parameter settings for \model{} models.

\begin{table}[H]
\resizebox{\textwidth}{!}{
\begin{tabular}{llllllllllllllll}
\toprule
 &  & \multicolumn{4}{c}{Encoder} & \multicolumn{4}{c}{Global Latent  Transf.} & \multicolumn{4}{c}{Decoder} & \multicolumn{2}{c}{Cross-Attn.} \\
 
Model &  & $l_\mathcal{E}$ & \#heads & $h_\mathcal{E}$ & \#Params & $l_\mathcal{G}$ & \#heads & $h_\mathcal{G}$ & \#Params & $l_\mathcal{D}$ & \#heads & $h_\mathcal{D}$ & \#Params & \#heads & k \\
\midrule
\textbf{400M} &  & 1 & 12 & 768 & 7M & 24 & 10 & 1280 & 470M & 7 & 12 & 768 & 50M & 10 & 2 \\
\textbf{1B} &  & 1 & 16 & 1024 & 12M & 25 & 16 & 2048 & 1B & 9 & 16 & 1024 & 113M & 16 & 2 \\
\textbf{2B} &  & 1 & 16 & 1024 & 12M & 26 & 20 & 2560 & 2B & 9 & 16 & 1024 & 113M & 16 & 3 \\
\textbf{4B} &  & 1 & 16 & 1024 & 12M  & 36 & 24 & 3072 & 4.1B & 9 & 16 & 1024 & 113M & 16 & 3 \\
\textbf{8B} &  & 1 & 20 & 1280 & 20M & 32 & 32 & 4096 & 6.4B & 6 & 20 & 1280 & 120M & 20 & 4 \\
\bottomrule
\end{tabular}}
\caption{Architectural hyper-parameters for different \model{} model sizes that we train for \flop{}-controlled experiments described in this paper.
}
\label{tab:arch}
\end{table}

\section{FLOPs Equations}
\label{section:flops-appendix}

Here, we provide the equations used for \flop{} computation for the forward-pass of transformer and \model{} models based on \cite{hoffmann2022training,kaplan2020scaling,casson2023flops}. We assume that the backward pass uses twice as much \flop{}s as the forward pass.

\begin{table}[H]
\centering
\begin{tabular}{@{}ll@{}}
\toprule
Operation & \flop{}s per token/byte  \\ \midrule
Attention $(l, h_k, n_{heads}, m)$ & $4 \times l \times h_k \times n_{heads} \times \frac{m+1}{2}$  \\
QKVO $(l, h, r)$ & $(r \times 2 + 2) \times 2 \times l \times h^2$  \\
Feed-forward $(l, h, d_{ff})$ & $2 \times l \times 2 \times h \times d_{ff}h$  \\
De-Embedding $(h, V)$ & $2 \times h \times |V|$ \\
Cross-Attention $(l, h_k, n_{heads}, p, r)$ &  \text{Attention}$(l, h_k, n_{heads}, p)$ + \text{QKVO}$(l, h_k \times n_{heads}, r)$ \\
\bottomrule
\end{tabular}
\caption{\flop{}s for operations used in transformer and \model{} models. $l$ corresponds to layers, $h$ is the hidden dimension ($h_k$ with $n_{heads}$ heads), $m$ is the context length, $d_{ff}=4$ is the feed-forward dimension multiplier, $p$ is the patch size, and $r$ is the ratio of queries to keys.}
\label{tab:transformer_flops}
\end{table}

For a transformer model with $l$ layers, hidden dimension $h$, context length $m$, $n_{heads}$ attention heads of dimension $h_k$, and a feed-forward multipler of $d_{ff}$, we compute \flop{}s as:
\begin{align}
\text{Transformer-FLOPs}(l, h, m, n_{heads}, h_k, d_{ff}, V) &= \text{Feed-forward}(l, h, d_{ff}) \\
&+ \text{QKVO}(l, h, r=1) \\
&+ \text{Attention}(l, h_k, n_{heads}, m) \\
&+ \text{De-Embedding}(h, V)
\end{align}

For \model{} models, we use the above-mentioned primitives together with the \flop{}s equation from Section \ref{section:flops} to compute total \flop{}s.

\section{Rolling Polynomial Hashing}
\label{appendix:rollpolyhash}
Given a byte $n$-gram $g_{i,n}=\{b_{i-n + 1},\ldots, b_i\}$, the rolling polynomial hash of $g_{i,n}$ is defined as:
\begin{align}
    \text{Hash}(g_{i,n}) &= \sum_{j=1}^n b_{i-j+1} a^{j-1} 
\end{align}
Where $a$ is chosen to be a 10-digit prime number.

\section{Frequency-based n-gram Embedddings}
\label{appendix:ngrams}
Prior to using hash n-gram embeddings in the final \model{} architecture, we also experimented with frequency-based n-gram embeddings. For each $n\in\{1,2,3,4,5,6,7,8\}$ there is an embedding matrix $E_{n}^{ngram}$ that contains the most frequent byte-grams for the given $n$.
Since it is intractable to store embeddings as $n$ grows, we only store embeddings for the most frequent $100,000$ byte-grams for each byte-gram.
If a particular position $i$ includes an $n$-gram present in the corresponding the embedding matrix, then this embedding is passed to the next step, encoder multi-headed cross-attention. If a byte-gram is infrequent and therefore not in the matrix, then its embedding is obtained from encoder hash embeddings instead.

Since frequency-based $n$-grams are limited by the vocabulary of the n-gram tables with infrequent n-grams not being represented at all, we subsequently moved to hash-based $n$-gram embeddings. See \autoref{tab:ngram_ablations_1b_all} for a comparison of hash and frequency based n-gram embeddings.

\begin{table}[H]
\centering
\resizebox{\textwidth}{!}{
\begin{tabular}{@{}lllllrrrr@{}}
\toprule
 & & &  &  &   \multicolumn{4}{c}{bpb} 
   \\
\cmidrule(lr){6-9} 
Hash Ngram Sizes & Per Hash Ngram Vocab & Ngram Sizes & Per Ngram Vocab & Total Vocab &  Wikipedia &    CC &  Github &  Train Dist \\

\midrule
               - &                  - &           - &               - &           - &      0.892 & 0.867 &   0.506 &       0.850 \\
           6,7,8 &                  50k &       6,7,8 &             50k &        300k &      0.878 & 0.860 &   0.497 &       0.843 \\
           6,7,8 &                 100k &           - &               - &        300k &      0.873 & 0.860 &   0.499 &       0.842 \\
           6,7,8 &                 100k &       6,7,8 &            100k &        600k &      0.868 & 0.857 &   0.494 &       0.839 \\
           6,7,8 &                 200k &           - &               - &        600k &      0.862 & 0.856 &   0.492 &       0.838 \\
           3,4,5 &                  50k &       3,4,5 &             50k &        300k &      0.862 & 0.856 &   0.491 &       0.837 \\
           3,4,5 &                 100k &           - &               - &        300k &      0.859 & 0.855 &   0.491 &       0.837 \\
           6,7,8 &                 200k &       6,7,8 &            200k &          1M &      0.861 & 0.855 &   0.491 &       0.837 \\
           6,7,8 &                 400k &           - &               - &          1M &      0.855 & 0.853 &   0.491 &       0.834 \\
     3,4,5,6,7,8 &                  50k & 3,4,5,6,7,8 &             50k &        600k &      0.855 & 0.853 &   0.488 &       0.834 \\
           3,4,5 &                 100k &       3,4,5 &            100k &        600k &      0.851 & 0.853 &   0.486 &       0.834 \\
           3,4,5 &                 200k &           - &               - &        600k &      0.850 & 0.852 &   0.485 &       0.833 \\
     3,4,5,6,7,8 &                 100k &           - &               - &        600k &      0.850 & 0.852 &   0.486 &       0.833 \\
           3,4,5 &                 400k &           - &               - &          1M &      0.844 & 0.851 &   0.483 &       0.832 \\
           3,4,5 &                 200k &       3,4,5 &            200k &          1M &      0.843 & 0.850 &   0.482 &       0.830 \\
     3,4,5,6,7,8 &                 100k & 3,4,5,6,7,8 &            100k &          1M &      0.844 & 0.850 &   0.482 &       0.830 \\
     3,4,5,6,7,8 &                 200k &           - &               - &          1M &      0.840 & 0.849 &   0.481 &       0.830 \\
     3,4,5,6,7,8 &                 200k & 3,4,5,6,7,8 &            200k &          2M &      \textbf{0.833} & \textbf{0.846} &   \textbf{0.478} &       \textbf{0.826} \\
     3,4,5,6,7,8 &                 400k &           - &               - &          2M &      \textbf{0.831} & \textbf{0.846} &   \textbf{0.478} &       \textbf{0.826} \\

\bottomrule
\end{tabular}
}
\caption{Ablations on the use of frequency-based as well as hash-based n-gram embedding tables for a 1B \model{} model trained on 100B bytes. }
\label{tab:ngram_ablations_1b_all}
\end{table}

\section{Entropy Patching Example from MMLU}
\label{sec:mmlu-appendix}

\begin{figure}[H]
\centering
\includegraphics[width=\textwidth]{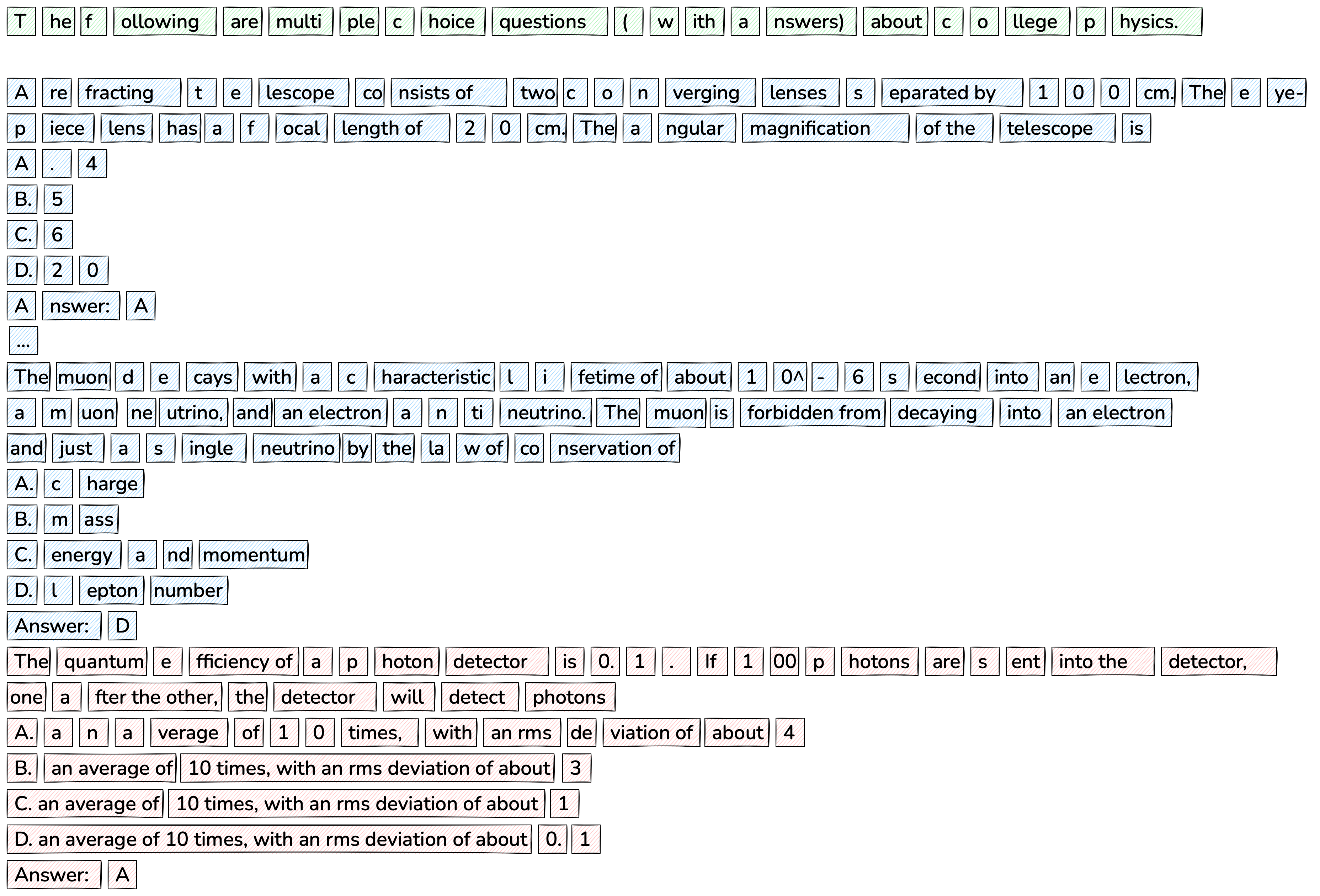}
\caption{An example of default entropy-based patching with global threshold during inference on \abr{mmlu}. Green denotes the prompt, Blue denotes the few-shot examples, and red denotes the question to be answered. Note that the size of the patches for the repeated phrases in the answer choices is much larger, which means that the global model is invoked significantly fewer times than its tokenizer-based counterpart, with this inference patching scheme.}
\label{fig:mmlu_patching}
\end{figure}

We illustrate how a few-shot example from a downstream task i.e. MMLU \citep{hendrycks2020measuring}, is patched using an entropy-model trained for use with \model{} models in Figure \ref{fig:mmlu_patching}. Directly using the entropy model with the full-context window causes repetitive patterns to be heavily patched. For example, ``10 times, with an rms deviation of about'' in the MMLU query is patched frequently the first time it is encountered, but is part of very large patches the next three times, which, although inference efficient, maybe undesirable for reasoning. One method that we use to avoid such a ``entropy'' drift is by resetting the entropy context with new lines and using a approximate monotonicity constraint (see Section \ref{section:entropy-context}).

\end{document}